\documentclass{article}

\PassOptionsToPackage{numbers, compress, comma, square}{natbib}
\usepackage[preprint]{neurips_2023}

\usepackage[utf8]{inputenc} 
\usepackage[T1]{fontenc}    
\usepackage[pagebackref]{hyperref}       
\usepackage{url}            
\usepackage{booktabs}       
\usepackage{amsfonts}       
\usepackage{nicefrac}       
\usepackage{microtype}      
\usepackage{xcolor}         

\usepackage{amsmath,amsfonts,bm}

\def\1{\bm{1}}

\def\vzero{{\bm{0}}}

\def\va{{\bm{a}}}

\def\vx{{\bm{x}}}

\def\vz{{\bm{z}}}

\def\mI{{\bm{I}}}

\def\mM{{\bm{M}}}

\def\mX{{\bm{X}}}

\DeclareMathAlphabet{\mathsfit}{\encodingdefault}{\sfdefault}{m}{sl}
\SetMathAlphabet{\mathsfit}{bold}{\encodingdefault}{\sfdefault}{bx}{n}

\def\sI{{\mathbb{I}}}

\def\sR{{\mathbb{R}}}

\usepackage[capitalise]{cleveref}
\usepackage{multirow}
\usepackage[ruled]{algorithm2e}
\usepackage{lipsum}  
\usepackage{scalerel}  
\usepackage{xcolor}  
\usepackage{array}  
\usepackage{dirtytalk}
\usepackage[font=footnotesize,skip=3pt,subrefformat=parens]{subcaption}
\usepackage{makecell}
\usepackage{soul}
\usepackage{subcaption}
\usepackage{graphicx}
\usepackage{wrapfig}
\usepackage[export]{adjustbox}

\usepackage{enumitem}
\usepackage{xspace}
\usepackage{subcaption}
\usepackage[font=small]{caption}
\usepackage[symbol]{footmisc}

\newcommand{\donut}{\protect\scalerel*{\includegraphics{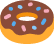}}{X}}
\newcommand{\ours}{DoughNet}

\crefname{section}{Sec.}{Secs.}
\crefname{figure}{Fig.}{Figs.}
\crefname{table}{Tab.}{Tabs.}

\title{DoughNet: A Visual Predictive Model for \\Topological Manipulation of Deformable Objects} 

\author{
Dominik Bauer $^1$ \quad Zhenjia Xu $^{1,2}$ \quad Shuran Song $^{1,2}$ \\
$^1$ Columbia University \quad $^2$ Stanford University \\
Corresponding author: \texttt{dominik.bauer@columbia.edu}\\
\href{https://dough-net.github.io}{https://dough-net.github.io}\vspace{-3mm}
}

\begin{document}

\maketitle
\begin{abstract}
Manipulation of elastoplastic objects like dough often involves topological changes such as splitting and merging. The ability to accurately predict these topological changes that a specific action might incur is critical for planning interactions with elastoplastic objects. We present DoughNet, a Transformer-based architecture for handling these challenges, consisting of two components. First, a denoising autoencoder represents deformable objects of varying topology as sets of latent codes. Second, a visual predictive model performs autoregressive set prediction to determine long-horizon geometrical deformation and topological changes purely in latent space. Given a partial initial state and desired manipulation trajectories, it infers all resulting object geometries and topologies at each step. DoughNet thereby allows to plan robotic manipulation; selecting a suited tool, its pose and opening width to recreate robot- or human-made goals. Our experiments in simulated and real environments show that DoughNet is able to significantly outperform related approaches that consider deformation only as geometrical change.
\end{abstract}

\section{Introduction}
\label{sec:intro}

From kneading to cutting, our interactions with elastoplastic objects such as dough or clay constantly involve actions that, beyond their overall geometry, manipulate the topology of objects.  
For example, changing the number of components allows to portion a roll of dough. By increasing its genus, we may form a doughnut from it. 
Different end-effector~(EE) geometries applying the same action to the same object, however, may yield drastically different topologies. 

\begin{figure}[!t]
    \centering
    \includegraphics[width=\textwidth]{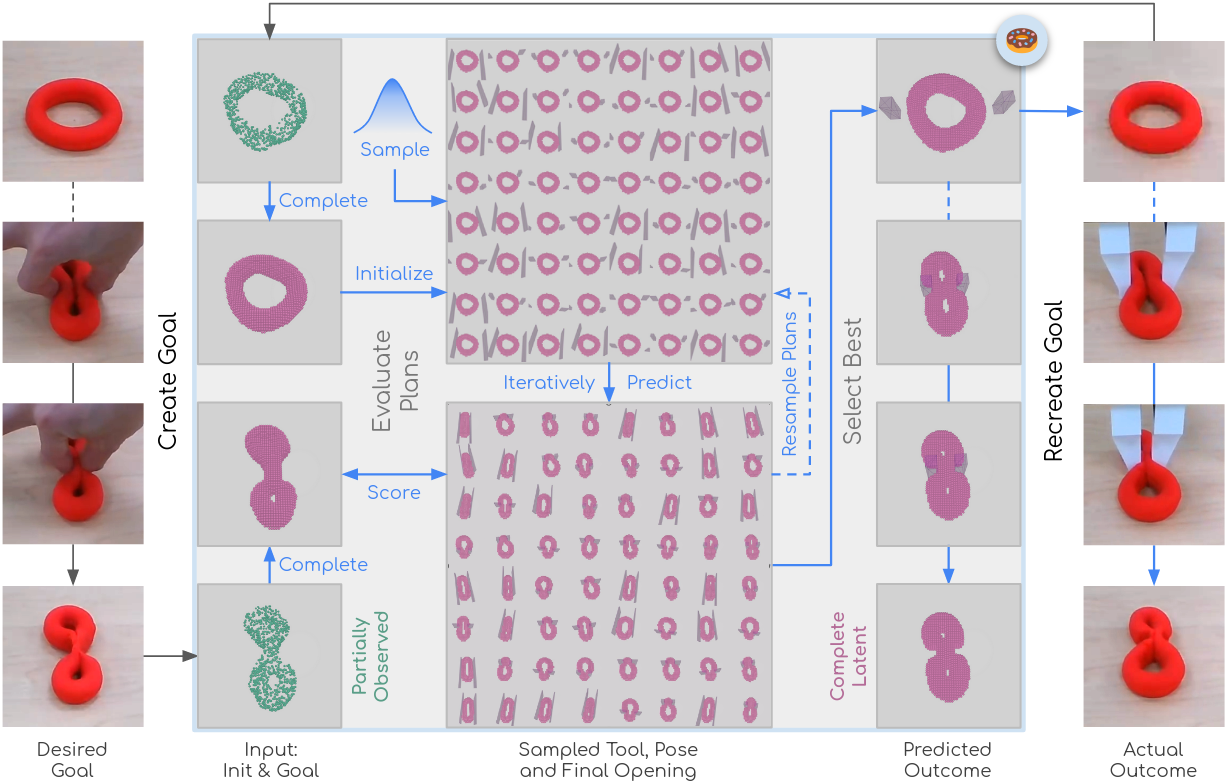}
    \caption{\textbf{Topological Manipulation.} Given partial point clouds of the initial and the goal state, \ours~predicts and scores the outcome of sampled plans. Executing the best found plan successfully recreates the goal.}
    \label{fig:teaser}
\end{figure}

In this paper, we focus on this challenging task of topological manipulation with a learned visual predictive model, illustrated in~\cref{fig:teaser}. Specifically, we infer the sequence of geometrical \textit{and} topological changes of an elastoplastic object -- given a single RGB-D observation of its initial geometry, EE geometry, and planned actions. Such a visual predictive model equips robots with the ability to plan their actions and choice of EE tools for achieving diverse topological manipulation goals, while still forming the desired object geometries.

Most prior works on elastoplastic manipulation represent objects by a single global geometry~\cite{shi2022robocraft,shi2023robocook,li2023dexdeform,qi2022learning}, or rely on additional post processing~\cite{hu2018mpm,heiden2021disect, lin2022pasta} to handle a specific topological change (i.e., splitting into two components). As a result, they are unable to predict more general forms of topological change. 
We conjecture that the exclusive focus on object geometry prevents a deeper investigation of topological manipulation. In the example of doughnut forming, while the geometry may be very similar, a roll with touching ends insufficently solves the task. Rather, they must also stay \textit{dynamically} connected when the object is moved. Determining these dynamic changes is, however, non-trivial. In the real world, conducting such a \say{topology check} destroys the geometrical state. In simulation, particle-based approaches~\cite{macklin2014pbd,hu2018mpm,hu2019taichi} may implicitly represent topology but the information is not readily available; mesh-based approaches~\cite{bathe2006finite,coumans2016pybullet,heiden2021disect} have explicit access to the topology but no trivial way to change it.

Aiming at a general solution to these challenges, we present \textbf{\ours}, a visual predictive model that infers objects' geometrical deformation and topological changes by jointly reasoning about objects as well as the EE geometry and its action trajectories. 
\ours~ consists of two main parts: First, a \textbf{denoising autoencoder} that represents geometries of different topological connectivity as sets of latent codes. These codes may be decoded into occupancy maps for connected components, for each of which we predict the genus.
Second, a \textbf{topology-aware dynamics model} learns the encoded objects' geometrical deformation and topological changes in an autoregressive set prediction task. Multi-step prediction is performed completely in the learned latent space, with both parts trained exclusively on simulated data. 
To generate training data with ground-truth topological structure, we propose a set of topological-checking operations for particle-based simulation. They enable us to reliably determine \textit{dynamic connectivity} between (merging, splitting) and within components (self-merging). 
Given a partial observation (from a single \mbox{RGB-D} camera) and a desired manipulation trajectory, \ours~allows to determine all resulting geometries and topologies, which we exploit for goal-directed topological manipulation.

 In summary, our main contribution is \ours, a visual predictive model for the novel task of topological manipulation of elastoplastic objects. It is enabled by 1) a topology-aware dynamics model that jointly reasons about objects' geometrical deformation and topological changes under different physical interactions, and 2) a synthetic data generator that yields volumetric geometry (as particles and meshes) and topological structure (as topology graphs and labels) for training the dynamics model.  
 Our experiments in simulated and real robotic environments show that \ours~outperforms previous approaches for predicting manipulation of deformables, especially in long-horizon predictions that facilitate the planning of robotic interactions.


\section{Related Work}
\label{sec:related}
We compare related predictive models for deformable objects along the axes of time, as well as geometry and topology.

\textbf{Temporal Abstraction.} While simulation step sizes may critically affect predictions, learning-based methods may explore different temporal abstractions.

One end of the spectrum is to consider whether a skill will eventually achieve the desired goal over long time horizons~\cite{lin2022pasta,lin2022diffskill,you2023make}. While this abstraction prevents error propagation of incremental prediction, it requires predicting high-DoF deformation in a single step. Hence, differentiable simulation may be additionally leveraged to improve upon the initial skill-based predictions~\cite{li2023dexdeform,you2023make}. 

On the other end of the temporal spectrum, Li et al.~\cite{li2019dpinet,li2020vgpl} train a GNN with data from particle-based simulation, predicting the corresponding particle graph's deformation in small increments. Shi et al.~\cite{shi2022robocraft,shi2023robocook} follow a similar GNN-based approach and incrementally predict manipulation using different tool geometries. The MPM simulation approach~\cite{sulsky1994particle} is the template for the NeRF-based~\cite{mildenhall2021nerf} dynamics model of \cite{li2022pac}.
Our approach yields more consistent shapes over longer manipulation horizons than incremental point-based predictions by learning a direct mapping between latent shapes at different time steps.


\textbf{Spatial Abstraction and Topological Change.} Whereas most analytical methods use particle, grid or tetrahedral representations, learned predictive models may be built upon a range of explicit and implicit shape representations.

Considering depth observations as input, for example, a common representation choice are point clouds and derived neighborhood graphs~\cite{li2019dpinet,li2020vgpl,shi2022robocraft,shi2023robocook,le2023differentiable,matl2021deformable}. While points may diffuse over longer prediction horizons, the added regularization due to nearest-neighbor graphs or mesh-like constraints hinders (keeping track of) topological change. For the latter, Heiden et al.~\cite{heiden2021disect} demonstrate that splitting may still be achieved by carefully designed geometrical processing.

Prediction methods~\cite{driess2023learning,van2023learning,seo2023masked,lin2022pasta,bartsch2023sculptbot,li2023dexdeform} that build upon recent advances in neural implicit representations~\cite{park2019deepsdf,peng2020convolutional,mildenhall2021nerf,zhang20233dshape2vecset}, in contrast, may exploit that the used occupancy or signed-distance fields naturally handle varying topologies. Commonly though, single objects~\cite{wi2022virdo,wi2022virdopp,bartsch2023sculptbot} or whole scenes~\cite{hafner2019dreamer,shen2022acid,driess2023learning} are represented by a global implicit shape. This may require the number of objects to remain unchanged or does not allow to easily determine if such a change indeed happened. For example, objects may be extracted in an initial step and embedded separately to discern them~\cite{lin2022pasta,driess2023learning}.
Inspired by Cheng et al.~\cite{cheng2021maskformer}, our method dynamically deals with varying connectivity -- and thereby topology -- during manipulation and makes these changes explicitly observable in distinct occupancy maps. No additional postprocessing of the predicted shapes is required.

\section{Method}
\label{sec:method}

\begin{figure*}[t]
    \centering
    \includegraphics[width=\linewidth]{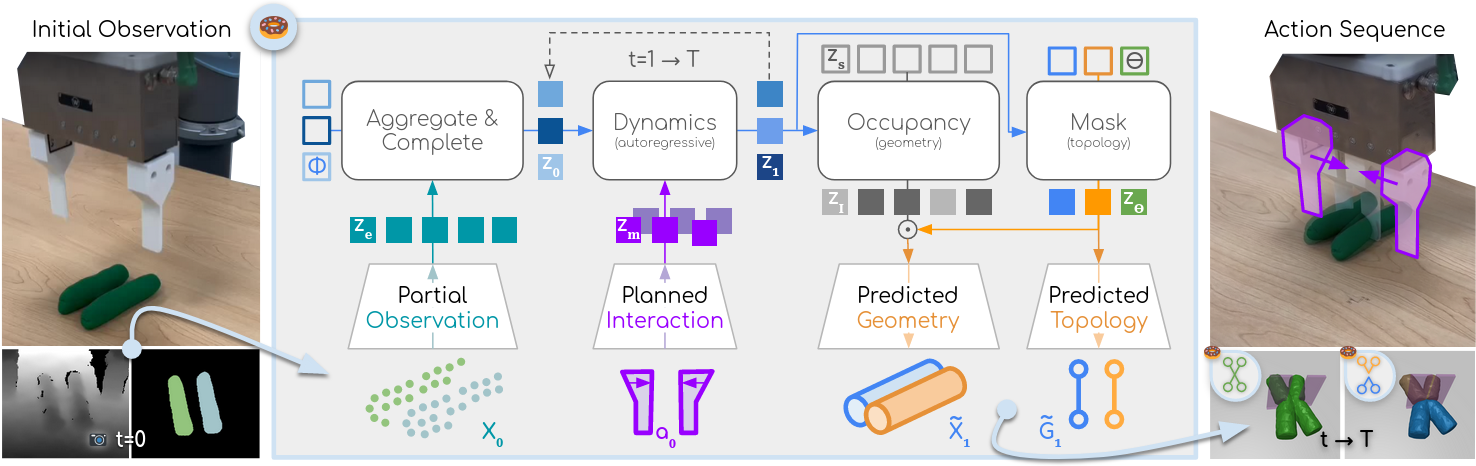}
    \caption{\textbf{\ours~Pipeline.} We encode the initial partial observation $X_0$ to a set of latent codes $[\vz_0]$ using a learned geometry embedding~$\Phi$. The given interaction $a_0$ yields the next latent codes $[\vz_1]$, which serve as input in subsequent time steps~$t\rightarrow T$. The latent codes may be reconstructed into components using a learned topology embedding $\theta$. This allows \ours~to reconstruct the objects' geometry $\Tilde{X}_t$ at sample locations~$[\vz_s]$. In addition, we may extract their topology $\Tilde{G}_t$ from the per-component latents $[\vz_\theta]$.}
    \label{fig:method}
    \vspace{-3mm}
\end{figure*}

Prior works commonly focus on the \textit{geometrical} changes that result from manipulation of deformable objects. However, few works~\cite{heiden2021disect,lin2022pasta} consider the \textit{topological} changes that occur when deformation leads to splitting or merging of objects. 
We argue that a joint consideration of both, geometry and topology, is important as it allows to minimize unwanted geometrical deformation when trying to achieve a topological goal (and vice-versa). 
For this joint reasoning, \ours~(presented in~\cref{fig:method}) is designed to perform two nested learning tasks, namely:
\begin{itemize}
    \item \textbf{Learning a latent representation} of deformable shapes with the encoder and decoder network, \cref{sec:met_encoder,sec:met_decoder}. 
    \item \textbf{Dynamics prediction} in the learned latent space, that infers objects' deformation and topological changes under different interactions, ~\cref{sec:met_prediction}.
\end{itemize}
Tying these learning objectives together, we describe the losses and training procedure of our method in~\cref{sec:met_loss}. While the autoencoder is supervised using a permutation-invariant reconstruction loss, the predictive model solely considers its prediction error in the latent shape space.

\subsection{Shape Encoder}\label{sec:met_encoder}

We hypothesize that repeated decoding, resampling and encoding would accumulate error and thus should be avoided. We therefore want to predict directly in latent space. As illustrated in~\cref{fig:method}, we achieve such a latent representation of the observed points $\mX \in \sR^{\text{N}\times(3+1)}$ by a set of latent codes $[\vz] \in \sR^{257\times512}$ in two main steps, based on the architecture proposed by Zhang et al.~\cite{zhang20233dshape2vecset}.

\textbf{Embed Partial Observation.} To embed the observed point cloud, we apply the feature transformation from~\cite{zhang20233dshape2vecset}. Each such feature is concatenated with its location $\vx_{xyz}$, fed through a linear layer and appended with the one-hot encoded component label $l~\in~\sI^P$ to create the per-point features $[\vz_{e}]$.

\textbf{Spatial Aggregation.} A learned query set $[\bm{\phi}]$ of $256$ latent codes aggregates over these per-point features via cross attention~\cite{vaswani2017attention}, $[\vz_{agg}] = \mathrm{CrossAttn}([\bm{\phi}],[\vz_{e}])$. To add further global information, inspired by~\cite{zhao2023michelangelo}, we add an average-pooled latent code by $[\vz] = [[\vz_{agg}], \overline{[\vz_{agg}]}]$.
The spatial aggregation to the latent set $[\vz]$ allows us to work with point clouds of varying size and contains the quadratic memory complexity of the following attention-based components by producing a latent shape set of small and fixed number. 
To further process the latent shape, i.e., learning to denoise and complete the partial observation, we leverage a stack of self-attention layers and their ability to consider global dependencies between the latents. While such a module is part of the decoder in~\cite{zhang20233dshape2vecset}, we observe significantly better performance when used in the encoder. Thereby, it already serves as input to the predictive model as compared to \say{postprocessing} its predictions.

\subsection{Dynamics Model}\label{sec:met_prediction}
Given a set of latent codes~$[\vz^t]$ in time step $t$, our model in~\cref{fig:method} predicts a new set~$[\vz^{t+1}]$ in time step~$t+1$. Its prediction is conditioned on the manipulator's interaction, i.e., its shape in both time steps. We assume this representation to be more general than an action space that depends on a specific kinematic chain.

\textbf{Embed Manipulator Geometry.} The manipulator's geometry is represented by a set of points uniformly sampled on its surface, combined for two time steps to describe its action $\va^t=[\mM^t,\mM^{t+1}]$. We conjecture that embedding the manipulator analogous to the observation facilitates finding interacting regions, e.g., for determining collisions that are implicitly required for dynamics prediction. The latent representation of the interaction is thus given by $[\vz^{t}_m,\vz^{t+1}_m]$, using $0$ as component label.

\textbf{Dynamics Prediction.} The embedded points are concatenated, forming the context~$[\vz^{t,t+1}_m] \in \sR^{257\times1024}$ for prediction based on the current object shape~$[\vz^t]$. We use four stacks that alternate two self- and with one cross-attention blocks to yield the next object shape~$[\vz^{t+1}]$. In our pipeline, this component is autoregressively applied to its own outputs to predict the outcome of long-horizon manipulations, including topological change.

\subsection{Geometry Decoder and Components}\label{sec:met_decoder}

To complete the autoencoder, we require a shape decoder that extracts both, geometry and topology, from the latent shape $[\vz]$. We jointly represent this information in the form of a set of $K+1$ occupancy masks $P(p_k|\vz, \bm{s}_i)$ for samples $[\bm{s}_i] \in \sR^{Sx3}$ and components $[p_k]_{k \in K}$. An additional mask represents samples that lie outside all shapes in the scene (\textit{outliers}).

\textbf{Interpolate Latents.} To decode the latent shape information at each sample location, samples are first transformed by $\vz_f(\bm{s}_i)$, concatenated with $\bm{s}_i$ itself and fed through a linear layer to yield $[\vz_s]$. Note that the linear layer is different from the one used by the encoder since we have no label information for samples. Next, analogous to the spatial aggregation, we employ a cross attention layer to interpolate the latent codes $[\vz]$ at the sample locations by $[\vz_{I}] = \mathrm{CrossAttn}([\vz_s], [\vz])$.

\textbf{Segment Samples.} As shown in~\cref{fig:method}, we combine this local decoder with a component prediction branch that is inspired by 2D segmentation~\cite{cheng2021maskformer}; instead of binary 2D masks, however, we predict a set of 3D occupancy masks. Correspondingly, we use a learned query set $[\bm{\theta}]$ of $K+1$ latent codes to compute a global component representation $[\vz_{\bm{\theta}}] = \mathrm{CrossAttn}([\bm{\theta}], [\vz])$. 
The dot product of the globally and locally decoded shape information is interpreted as the logits of the component occupancy probability $P(p_k|\vz, \bm{s}_i)$ for component $p_k$ at sample location $\bm{s}_i$. We prescribe each sample to belong to exactly one component (or the outliers) and extract the label that maximizes $P$.

\textbf{Predict Topology.} We integrate topology prediction as a classification task into our pipeline. In contrast to previous work~\cite{chen2020topology}, we simplify this task by leveraging the \say{pre-segmented} components. Classifying each component latent $\vz_{\bm{\theta}}^i$ to either a class corresponding to its genus or an \textit{empty} class allows us to additionally obtain the number of components using the same network head. As a practical benefit, moreover, one does not need to query the occupancy for the latent geometry of the corresponding components.

\subsection{Training Procedure and Losses}\label{sec:met_loss}

Since the required ground-truth components are infeasible to obtain for real observations (i.e., determining a topological change without intervention and thus without destroying the state of the deformable object), we base our training on synthetic data described in~\cref{sec:dataset}. We assume that component meshes are available for each time step of the manipulation trajectories in the training set.

\textbf{Generate Synthetic Observations.} To generate realistic training data, we render depth images of these meshes from jittered camera poses. Sensor noise is added as a normally distributed per-pixel offset, by random pixel shifts and by removing pixels with large depth difference~\cite{denninger2023blenderproc,tolgyessy2021evaluation}. Rendering at a lower resolution and upscaling via bilinear interpolation introduces smoothing of the depth image, creating transitions between objects and the environment that further reduce the domain gap.

\textbf{Reconstruct and Complete the Observations.} The order of the $K$ occupancy masks is ambiguous; the component label has no semantic meaning since it only encodes distinct connectivity. Therefore, similar to Cheng et al.~\cite{cheng2021maskformer}, we compute the reconstruction loss using the matching $\mathcal{L}_{rec!} = \min_{K \in \mathcal{K}!} \mathcal{L}_{rec}(K, \mathcal{K}_{true})$, where $\mathcal{K}!$ is the set of all permutations of the $K$ predicted components and $\mathcal{K}_{true}$ their true permutation. An additional component for outliers is kept at the same index to facilitate its rejection. We use the Focal Loss \cite{lin2017focal} to score these permutations. 
This accounts for class imbalance, i.e., \textit{occupied versus empty space} in our case. The best-component permutation is also used in the cross-entropy loss supervising the topology classifier.

While the encoder is only given a partial and noisy observation, the targets are computed from the true shapes and labels by computing the signed distance with respect to the component meshes at the sample locations $[\bm{s}_i]$. Assuming non-intersecting meshes, a sample is assigned to the outliers if all signed distances are positive (outside all meshes) and to the component for which it is $\leq 0$ else.

\textbf{Predict the Next Shapes.} We want our prediction model to iteratively find the shapes that follow from the given interactions in the next time step, i.e., the latent codes that would decode to the complete ground-truth shape at the respective time step. 
Therefore, we directly supervise the predictive model in latent space by the Huber loss~\cite{huber1964robust} between the latent set of the prediction $\vz_{pred}$ and the corresponding encoder output $\vz$. 
We combine this loss for two time steps, namely $t,t+1$. In the first time step, the prediction $[\vz^t_{pred}]$ is computed from the encoded shape $[\vz^{t-1}]$. To reduce the exposure bias commonly observed in autoregressive models~\cite{bengio2015scheduled,schmidt2019generalization}, we use $[\vz^{t+1}_{pred}]$ computed from $[\vz^t_{pred}]$ in the second time step. Thereby, the predictive model must learn to map latents from the encoder as well as itself to the same shape. The topology classifier is also trained on latents from both sources.
We find training to be more stable if we pretrain the autoencoder using the reconstruction loss and, in a second stage, freeze it and only train the predictive model using the latent loss.

\section{Topological Manipulation Environment}
\label{sec:dataset}
Synthetic data generated with an MPM-based simulator is used for training. The simulation setup is presented in \cref{sec:simulation_setup}. Ground-truth information is extracted as training supervision, including both geometry and topology structure via a set of topological-checking operations (\cref{sec:topological_checking}).
Finally, the model trained with synthetic data is directly evaluated on a real-world platform (\cref{sec:real_env}). 

\begin{figure}[t]
    \centering
    \includegraphics[width=\textwidth]{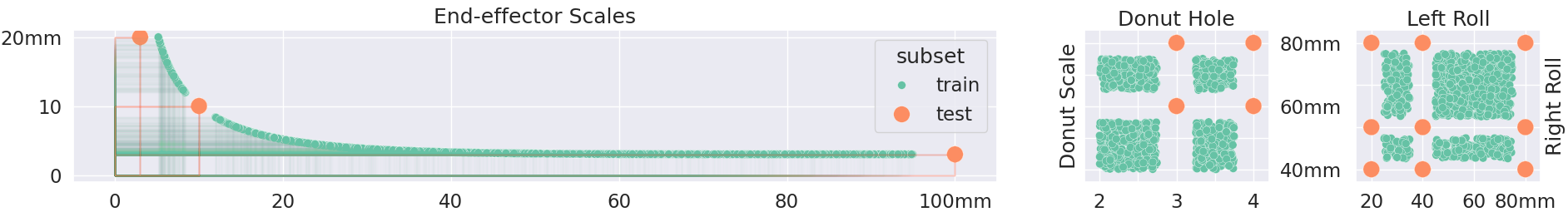}
    \caption{ \footnotesize \textbf{Data Distribution.} The scales of the objects and EE geometry, where test samples are held-out from regions inside and outside the training boundaries.}
    \label{fig:dataset-generation}
\end{figure}

\subsection{Simulation Setup}
\label{sec:simulation_setup}
The simulation environment consists of one or two objects made of soft material and an end-effector (i.e., parallel-jaw gripper) with varying finger geometry. The initially opened gripper closes gradually until reaching the minimal opening width at a varying in-plane pose. Within the process, full object particle and manipulator mesh states are recorded at constant intervals (\say{keyframes}).

The soft material is represented using an elastoplastic continuum model simulated with MLS-MPM~\cite{hu2018mpm} and von Mises yield criterion. MPM-based methods naturally support arbitrary deformation and topology change. The rigid gripper fingers are expressed as time-varying signed distance fields (SDFs), derived from external meshes. Contact between soft and rigid materials is modeled through the computation of surface normals of the SDFs.

We create simulation environments that are visually simple scenes, yet undergo complex topological changes in terms of the number of components ($c$) and genera ($g$). 
Scenes with two \say{rolls of dough} side-by-side (initially $c=2$) may merge (to $c=1$) and split (to $c=2, 3$ or $4$). Scenes with a \say{doughnut} (initially $g=\{1\}$) may self-merge (to $g=\{0\}$ or $\{2\}$) and split (to any combination of $g=\{0,1\}$), where the resulting topology is highly sensitive on the EE's geometry and pose. For each type of scene, we randomly sample 1000 training and 100 testing instances for a total of 2000 and 200 scenes. Six faulty test scenes are discarded. Each scene is subsampled to 13 frames for larger displacements. %

We vary the end effector geometry from thin \say{scissors} to a wide \say{vise}, starting from a random in-plane pose and closing its fingers to a minimal distance. We hold out a range of scales around the test geometries, as shown in~\cref{fig:dataset-generation}.
For the EE and scenes, the sampled parameter regions for training and testing are disjoint and include cases requiring interpolation and extrapolation.

\subsection{Topological Check}
\label{sec:topological_checking}

Topological connectivity is implicit in MPM-based simulation; it operates on a set of particles and a corresponding grid, not on well-separated objects. Therefore, (self-)merging and splitting are ill-defined from particle locations alone. 
For example, distance-based connectivity cannot distinguish two cylinders lying next to each other (in contact) from ones that have been squeezed (merged).
Instead, we define their connectivity by the counterfactual argument that \say{they would not stay in contact \textit{when moved apart}, if they had not merged or had split}.

\begin{wrapfigure}{r}{0.45\textwidth}
\vspace{-3mm}
   \centering
   \includegraphics[width=0.45\textwidth]{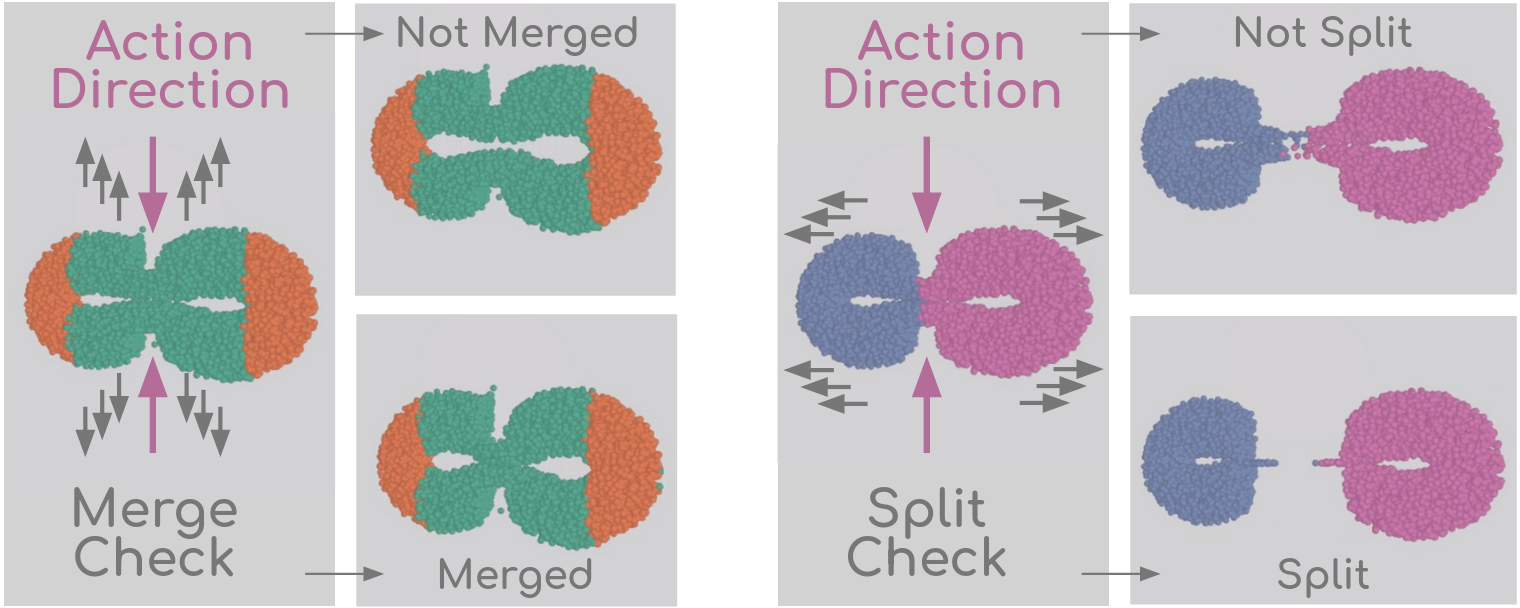}
    \caption{\textbf{Topology Check.} Left: Two components (same or different object) are considered merged \textit{iff} they remain connected after opposite velocities are applied. Right: A component is considered split into two \textit{iff} they have no connection after checking.}
    \label{fig:topological_checking}
\end{wrapfigure}

To this end, we propose a physically plausible topology annotation by focusing on connectivity changes induced by the EE.
We derive the object regions between the EE fingers where a merge or split may occur  from the kNN graph of particles at rest and their current locations. 
As shown in \cref{fig:topological_checking}, our simulation-based topology check involves adjusting particle velocities to move these object regions opposite to the end effector's closing direction (merge) or orthogonal to it (split).
If the distance between these regions after perturbation by the checking action falls below (merge) or above (split) a threshold, this topological change is recorded in both the particle-connectivity graph and the scene-topology graph.
The latter is the objects' initial topology in terms of a simple planar graph. Each recorded change alters this graph using merge, split, and remove graph operations. To support, e.g., complex self-merge events, our operations also take the particles surrounding the check regions into account. This scene-topology graph is easily evaluated in terms of the number of connected components and cycles, where the number of cycles is equivalent to the genus of the corresponding object.

While the particle representation lends itself to simulation, for further processing such as resampling the object geometry's surface or interior, a surface mesh representation is more flexible. 
Equipped with the dense labeling in the particle-connectivity graph, we can select all particles that belong to an individual component and compute its corresponding surface mesh. To this end, we compute the signed distance with respect to the each component's particles in a regular grid and apply Dual Contouring~\cite{ju2002dualcontouring,gibson1999surfacenets}, which is designed to maintain sharp features -- such as cuts in deformable objects. Our final processed dataset minimally consists only of color-coded component meshes and manipulator meshes with poses, as well as the scene-topology graph.

\section{Experiments}
\label{sec:experiments}

Going beyond predicting deformation of geometry, we show that \ours~is able to also accurately predict the resulting changes in connectivity as components merge and split during manipulation. We outline our experimental procedure, 
metrics and baselines in~\cref{sec:ex_metrics}. Results for synthetic and real-world robotic interactions are presented in~\cref{sec:ex_prediction,sec:real_env,sec:planning}. 

\textbf{Procedure and Metrics.}
\label{sec:ex_metrics}
Given a partial point cloud, the initial task is to \textit{reconstruct} the complete objects.
A \textit{single step} prediction observes the scene in every step and predicts one step into the future. For \textit{full sequence} prediction, only a single initial observation is given and methods need to generate subsequent future states from their own output. 
We measure performance using the following four metrics, computed per frame and reported as mean over all frames:
\begin{itemize}
    \item \textbf{Voxel IoU (VIoU)} compares the predicted and true scene occupancy (i.e., over all components). It thus only measures the geometrical quality. Occupancy is evaluated on a grid with the equivalent of a 1mm spacing, ignoring points below the plane or within the EE since they are assumed to be known.
    \item \textbf{Component IoU (CIoU)} compares the occupancy per true component. It shows both geometrical and topological quality. We report the mean over the best matching components' IoU. 
    \item \textbf{Accuracy of the Number of Components (AccC)} is the percentage for which methods are able to predict the correct number of components. 
    \item \textbf{Accuracy of the Genus (AccG)} is the equivalent metric for the genera of these components. For consistency, the permutation found for the CIoU metric is reused here to match the components. 
\end{itemize}

\textbf{Baselines.}
\label{sec:ex_baselines}
We compare \ours~to baselines that feature different considerations of (1)~geometry representation (particle point cloud or occupancy), (2)~topology (post-hoc static or dynamic check), and (3)~prediction (simulation- or learning-based):
\begin{itemize}
    \item \textbf{VPM} (Visual Predictive Model) is a variant of our approach that discards the predicted components and instead computes the topology from scratch. These results motivate the inclusion of topology into our joint prediction.
    \item \textbf{MLS-MPM}~\cite{hu2018mpm} is a predictive model using the MPM simulator and our initial geometry reconstruction. This algorithm would provide a performance oracle for the dynamics prediction. However, the simulation could be sensitive to initialization and less robust to partial and noisy observations.
    \item \textbf{RoboCook}~\cite{shi2023robocook} is a GNN-based predictive model of object deformation. Building upon a line of previous works~\cite{shi2022robocraft,li2020vgpl,li2019dpinet}, it is representative of learned models that closely follow the design of particle-based simulators. Our comparison therefore shows the benefit of reasoning on an object level.
\end{itemize}
Since ours is the only method to explicitly reason about topology, we furthermore augment these baselines by post-hoc topology checks:
\begin{itemize}
    \item \textbf{Static topology check} extracts the connected components from the nearest-neighbor graph~\cite{pearce2005cluster}. A mesh is computed for each, using the Euler characteristic to obtain its genus. This baseline illustrates the necessity of considering \textit{dynamic} connectivity; beyond using only static Euclidean distances.
    \item \textbf{Dynamic topology check} uses the same approach used for our synthetic data generation. As this requires repeated simulation, it is combined with the MLS-MPM baseline and highlights the computational burden of both.
\end{itemize}

\begin{table}[t]
\setlength{\tabcolsep}{2pt}
    \centering
    \begin{tabular}{l|cccc|cccc|r}
    \toprule
   & \multicolumn{4}{c|}{\textit{full sequence}}& \multicolumn{4}{c|}{\textit{at final time step}}&\\
              & VIoU$^\uparrow$ & CIoU$^\uparrow$ & AccC$^\uparrow$ & AccG$^\uparrow$ & VIoU$^\uparrow$ & CIoU$^\uparrow$ & AccC$^\uparrow$ & AccG$^\uparrow$ & fps$^\uparrow$ \\
        \hline
          \ours~\donut& \textbf{92.0}& \textbf{90.5}& \textbf{97.9} & 98.6 & \textbf{84.1}& 75.1& \textbf{92.3}& 90.3& \textbf{22.1} \\
        \hline
          VPM$^+$& \textbf{92.0}& 88.5& 92.8& 94.8 & 83.3& 70.7& 68.7& 79.7& 0.4 \\
          MLS-MPM~\cite{hu2018mpm}$^\ddagger$& 87.2& 86.2& 97.2& \textbf{98.9}& 82.0& \textbf{79.4}& 91.8& \textbf{95.6}& 0.2 \\
          RoboCook~\cite{shi2023robocook}$^+$& 60.8& 59.2& 91.5& 95.6& 44.1& 37.4& 64.1& 80.9& 4.5 \\
          \bottomrule
    \end{tabular}
    \vspace{2ex}
    \caption{\textbf{Performance on Simulated Data.} Results for all 194 test scenes (13 time steps each). $+$~with a static topological check, $\ddagger$~with our proposed dynamic topological check and ours~(\donut) with predicted topology. Frame rates (fps) are for geometry \textit{and} topology prediction, without exploiting parallel simulation or batched prediction, on a system equipped with an Intel Xeon Gold 6248 and an NVIDIA Geforce RTX 2080~Ti.}
    \label{tab:ex_main}
    \vspace{-3mm}
\end{table}

\begin{figure}[t]
    \includegraphics[width=\linewidth,trim={3mm 6mm 0 0},clip]{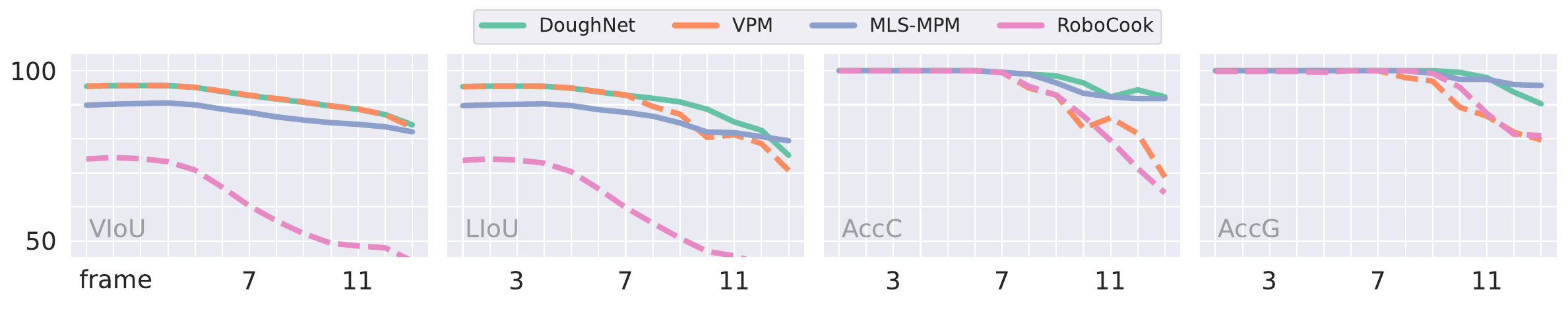}
    \caption{\textbf{Performance on Simulated Sequences}. Mean performance per frame; dashed for the static topology check~($^+$), and solid for dynamic~($^\ddagger$) and predicted~(\donut).}
    \label{fig:ex_main_time}
\end{figure}
\begin{figure}[!t]
    \includegraphics[width=\linewidth]{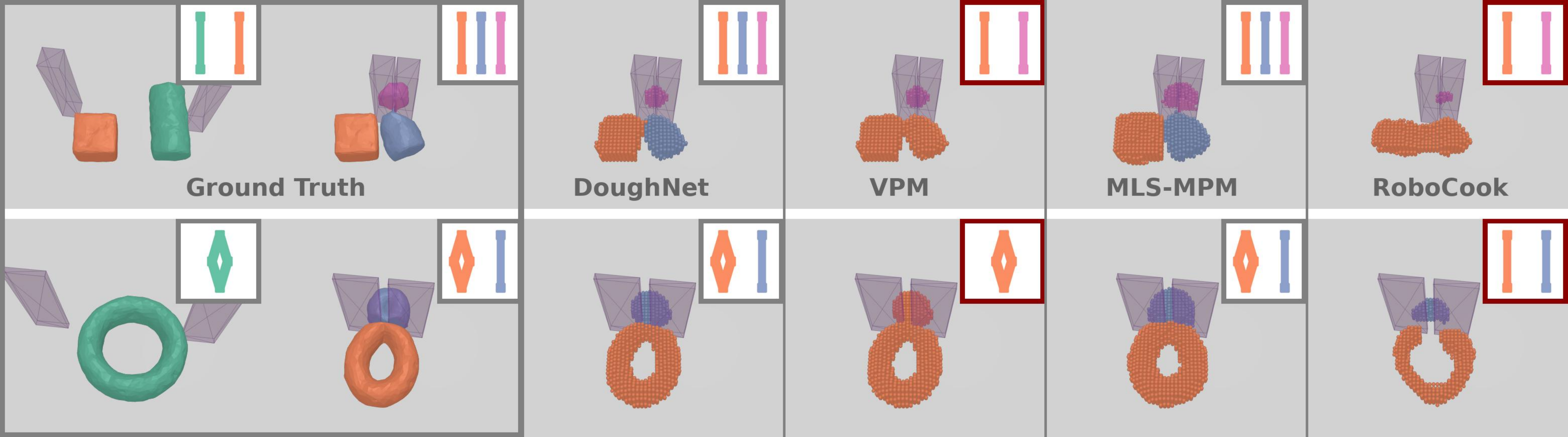}
    \caption{\textbf{Performance on Simulated Sequences}. Initial state and comparison at final time step $t=13$. Components are color-coded and the boxes in the top right corner of each sample indicate the scene-topology graph (its order has no meaning).}
    \label{fig:ex_main_quali}
\end{figure}

\subsection{Results}
\label{sec:ex_prediction}

As indicated in~\cref{tab:ex_main}, \ours~significantly outperforms the baselines in full sequence prediction, retraining high accuracy over a long prediction horizon. In~\cref{tab:ablation}, we compare the effect of key design decisions on prediction quality.

\textbf{Comparison to Post-hoc Topology Prediction.} 
The significant performance drop of using a static topology check for VPM illustrates that the effects of topological manipulation go beyond geometrical changes. Our approach, instead, learns to exploit the observations generated by the proposed dynamic check in simulation to jointly reason about such topological change, rather than relying on post-hoc clustering of predicted points. 

\textbf{Comparison to Simulation-based Prediction.} As shown in~\cref{tab:ex_main}, only in the final time step, MLS-MPM -- the simulation initialized with our reconstruction and using our proposed check as topology oracle -- performs better on the CIoU and AccG metrics, while still being outperformed by ours in terms of VIoU and AccC. This drop may be attributed to \ours~at times requiring additional steps to cleanly resolve a topological change in the per-component occupancy. Still, we achieve better or comparable results but magnitudes faster.

\textbf{Comparison to GNN-based Prediction.} While \ours~maintains high performance over subsequent predictions, RoboCook's performance deteriorates over time in~\cref{fig:ex_main_time}. It is further diminished by the downsampling to a sparse point cloud as input for the GNN, necessitating costly meshing for upsampling to a dense (yet still lossy) prediction again in each frame. DoughNet achieves a 31.3\% higher CIoU than RoboCook. This comparison supports our choice of an implicit representation that may be evaluated at any resolution, jointly handling geometrical and topological change without loss in quality.


\begin{table}[t]
    \setlength{\tabcolsep}{2pt}
    \centering
\begin{tabular}{l|cccc|cccc}
    \toprule
     & \multicolumn{4}{c|}{\textit{single step}}& \multicolumn{4}{c}{\textit{full sequence}}\\
             &    VIoU$^\uparrow$&CIoU$^\uparrow$& AccC$^\uparrow$&AccG$^\uparrow$& VIoU$^\uparrow$& CIoU$^\uparrow$& AccC$^\uparrow$&AccG$^\uparrow$\\ \hline
     \ours~\donut& 94.3& 93.0& 98.2& 98.8& \textbf{92.0}& \textbf{90.5}& \textbf{97.9}&\textbf{98.6}\\ \hline
 one-step prediction& 94.1& 92.7& 98.1& 98.8& 90.8& 89.0& 97.4&98.4\\
 explicit supervision& \textbf{94.8}& \textbf{93.7}& 98.0& 98.7& 76.3& 75.4& 91.7&93.1\\ 
 relative prediction& 
          94.4&93.2&\textbf{98.4}&\textbf{99.0}& 73.1& 74.2& 80.4&81.6\\
 complete observation& 77.0& 75.3& 95.5&98.1& 75.8& 73.4& 95.7&96.9\\
     \bottomrule
\end{tabular}
    \vspace{2ex}
    \caption{\textbf{Ablation Study on Design Decisions.} In each row, we replace one design decision with an alternative described in~\cref{sec:ex_prediction}. We note the significant challenge of long-horizon prediction of topological manipulation, as compared to single steps.
    }
    \vspace{-5mm}
    \label{tab:ablation}
\end{table}

\textbf{Effect of Training Multi-step Prediction.} Prior works~\cite{bengio2015scheduled,schmidt2019generalization} observe that autoregressive models, as ours, suffer from an \textit{exposure bias}; training on ground-truth inputs but using own predictions as input in subsequent steps at test time. Shi et al.~\cite{shi2023robocook} are able to reduce this effect by predicting two time steps during training. Following this observation, we predict one step from reconstruction of the true state and one step from the prediction of the next step. As compared to only training the first time step, there is a slight improved performance in the \textit{single step} setting that increases to a bit over 1\% on the IoU metrics when moving to \textit{full sequences}, as shown in~\cref{tab:ablation}. 

\begin{wrapfigure}{r}{0.55\textwidth}
    \vspace{-4mm}
    \centering
    \includegraphics[width=0.55\textwidth]{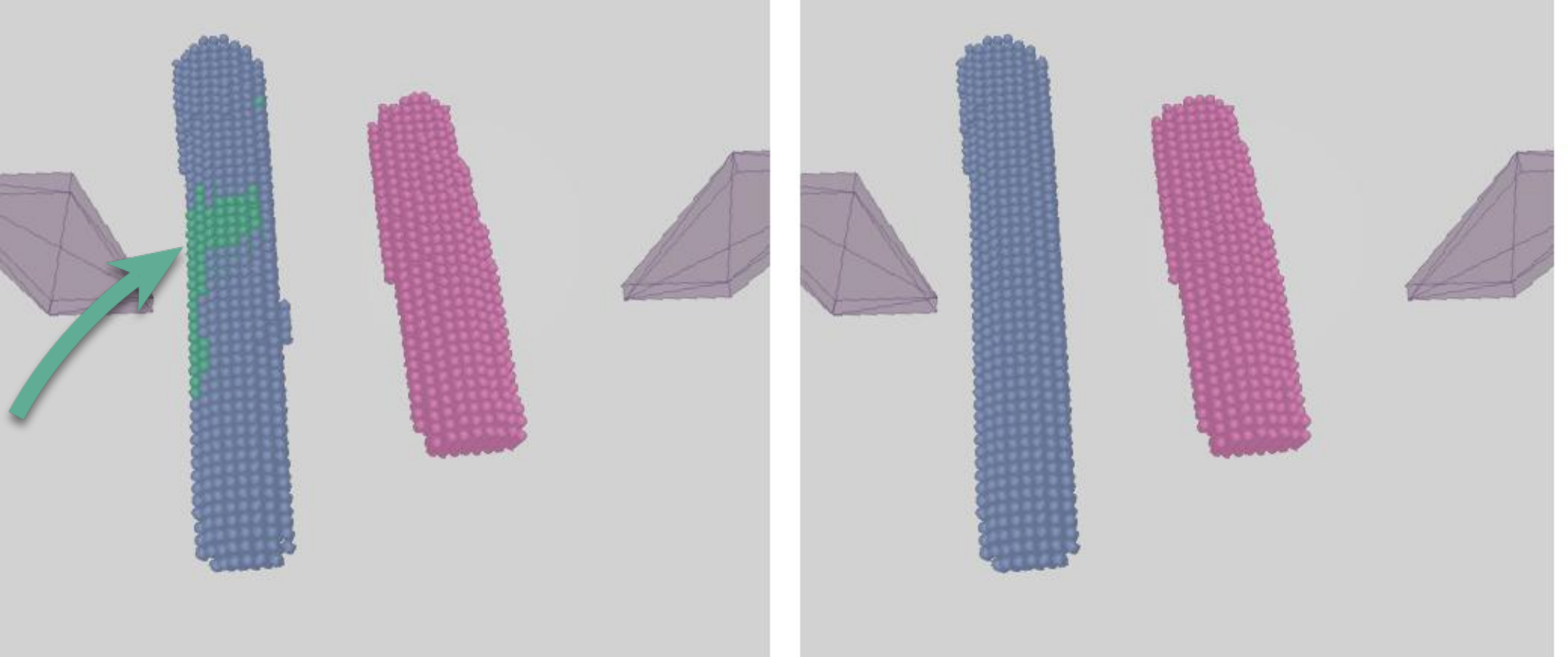}
    \caption{\footnotesize \textbf{Error Correction.} Initial reconstruction errors (left) are corrected by our predictive model in next steps (right).}
    \label{fig:correction}
    \vspace{-3mm}
\end{wrapfigure}

\textbf{Effect of Latent Space Supervision.} In RoboCook~\cite{shi2023robocook}, prediction is supervised by the reconstruction's distances to the ground truth -- a small error in latent space is not penalized during training as long as the reconstruction is similar. We hypothesize that this is an additional source for error accumulation in sequential prediction. Indeed, we find that directly supervising the prediction of the latents is more robust as this forces the predictive model to map to the space learned by the autoencoder, further avoiding degenerate reconstructions (see~\cref{fig:correction}). Quantiatively, we observe a significantly improved long-horizon performance using our approach, achieving about 15\% higher IoU scores in~\cref{tab:ablation}.

\textbf{Effect of Set-to-set Prediction} An approach in previous visual predictive models, perhaps inspired by a Lagragian view on dynamics, is to learn the relative change between states. Like any iterative procedure without error correction, we conjecture that this approach is prone to error accumulation. Rather, we directly predict the next state -- from one set of latents to the next. The results in~\cref{tab:ablation} quantify the benefit of this, specifically a 18.9\% VIoU increase and 17.5\% AccC increase compared to the \textit{relative} variant. Further supporting our decision, we observe that our predictive model is able to correct, e.g., initial reconstructions from real point-cloud observations in terms of topological errors, shown in~\cref{fig:correction}, and geometrical inaccuracies.

\textbf{Effect of Training on Partial Observations.} Rather than assuming a complete initial state (as for MLS-MPM~\cite{hu2018mpm}) or a multi-view reconstruction (as in~\cite{shi2023robocook}), we conjecture that a single partial observation is sufficient. To bridge the domain gap between synthetic and real point-cloud observations,  we introduce common artefacts of depth sensing, namely incompleteness, oversmoothing, and fringing at transitions to the background. Experiments on synthetic partial data in~\cref{tab:ablation} show an expected yet significant improvement using our data preparation pipeline as compared to training on complete observations. Importantly, as shown in~\cref{fig:ex_main_qualitative}, this robustness to partial input carries over to experiments on the robot and enables successful sim-to-real transfer. We want to highlight that, predicting from a single partial observation, we even achieve plausible reconstructions of components that are eventually hidden from direct observation.

\subsection{Real-world Evaluation}
\label{sec:real_env}

\begin{wrapfigure}{r}{0.4\textwidth}
    \vspace{-5mm}
    \includegraphics[width=0.4\textwidth]{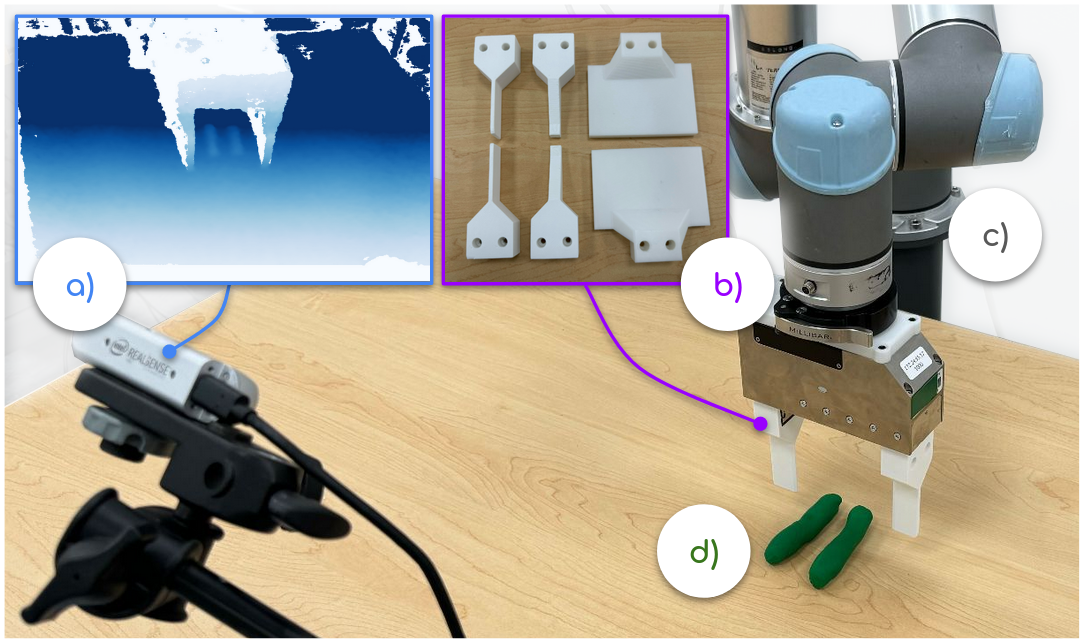}
    \caption{\footnotesize \textbf{Real-world Setup.} a) Intel RealSense D415 (and depth image), b) Weiss Robotics WSG50 (with three 3D-printed tools of held-out geometry), attached to c) a Universal Robots UR5 to manipulate d) molded Play-Doh objects.}
    \label{fig:real-setup}
    \vspace{-4mm}
\end{wrapfigure}

\begin{figure}[h]
    \centering
    \includegraphics[height=0.395\linewidth]{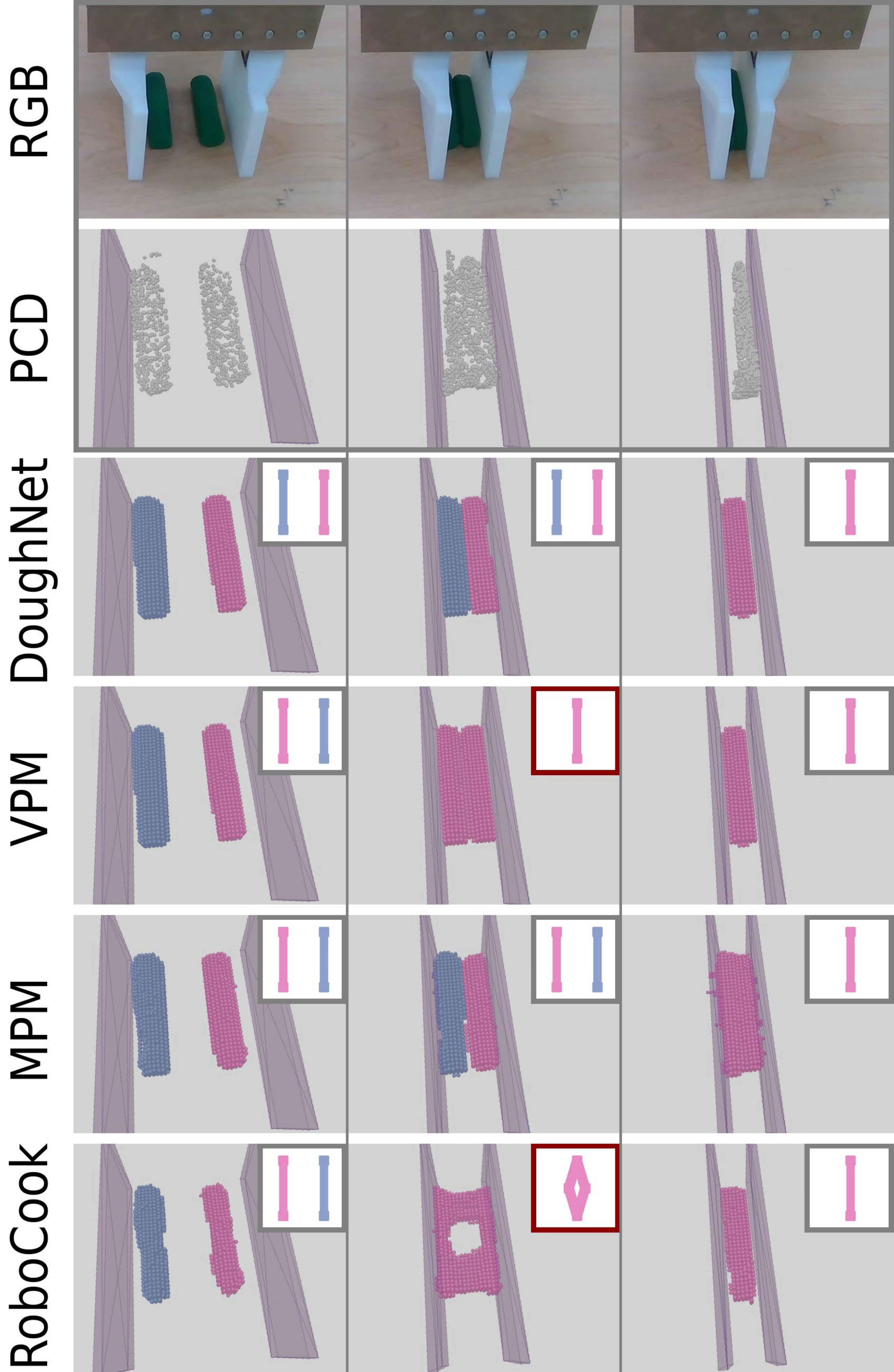}
    \includegraphics[height=0.395\linewidth]{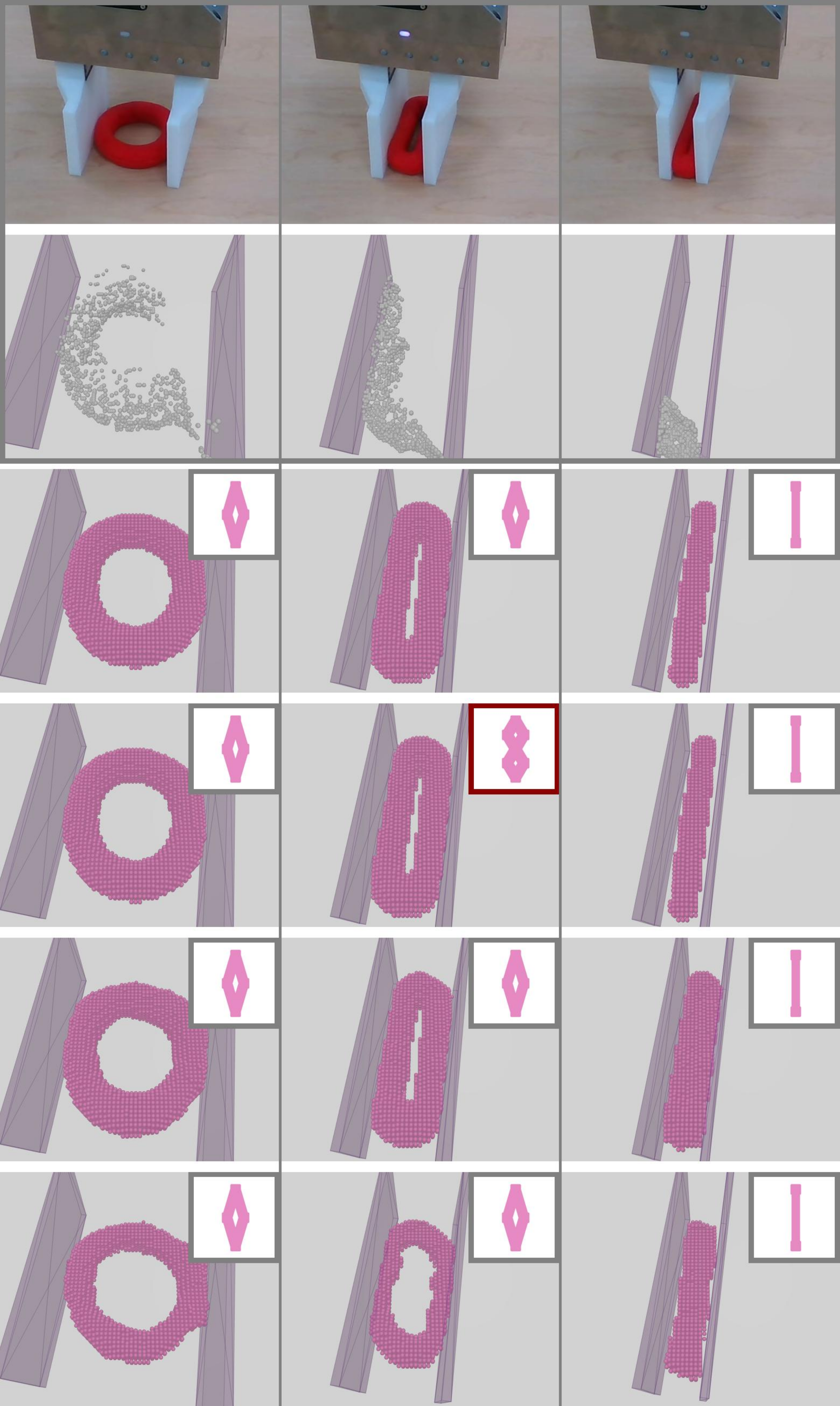}
    \includegraphics[height=0.395\linewidth]{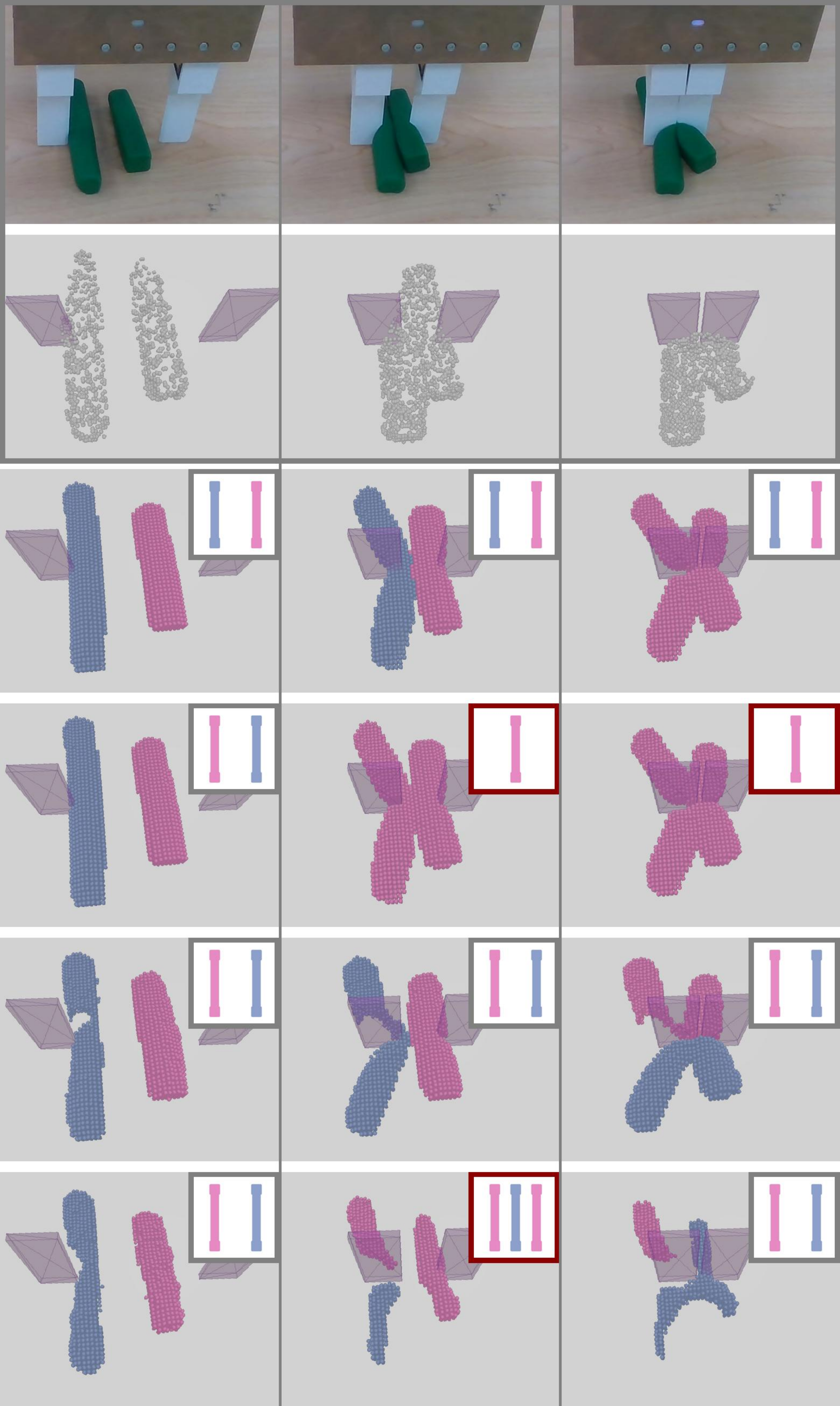}
    \includegraphics[height=0.395\linewidth]{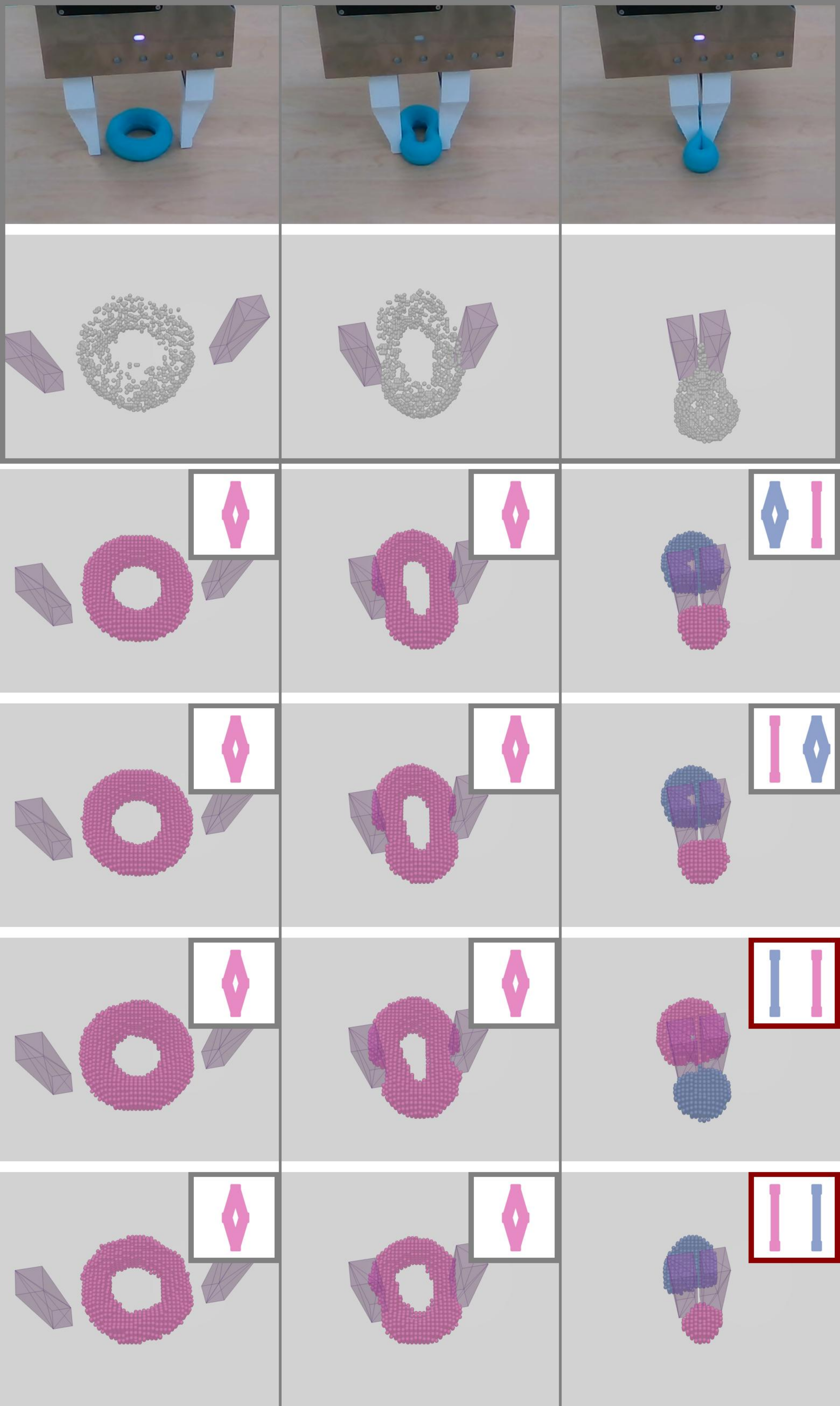}
    \caption{\textbf{Qualitative Comparison on Real Sequences}. Full-sequence predictions from an initial partial observation. Results are shown at time steps $t=\{5, 10, 15\}$. Erroneous topology predictions (white boxes, top right) feature a dark red border.}
    \label{fig:ex_main_qualitative}
\end{figure}

We further evaluate our model on a real-world platform as shown in \cref{fig:real-setup}. The scene setup and placement strategy follows the synthetic dataset. The setup is observed using a single camera, which provides the initial partial and noise-afflicted observation for our model. A total of 60 scenes (\textit{interactions}) of 15 frames each are recorded.
In a preprocessing step, we segment the objects of interest by color and crop the resulting point cloud by the known table plane. Statistical outlier removal and voxel downsampling reduce the gap between synthetic and real data. Furthermore, for two-roll scenes, PCA-based reorientation and centering along the median axis between the objects bring the observation to a (noise-afflicted) canonical frame, in which we can easily determine the initial components.
The qualitative results are shown in~\cref{fig:ex_main_qualitative}, demonstrating that our model trained on purely synthetic data is able to generate plausible deformation and topology changes from single RGB-D images in real robotics experiments.

\begin{figure}[h]
    \centering
    \includegraphics[width=\linewidth]{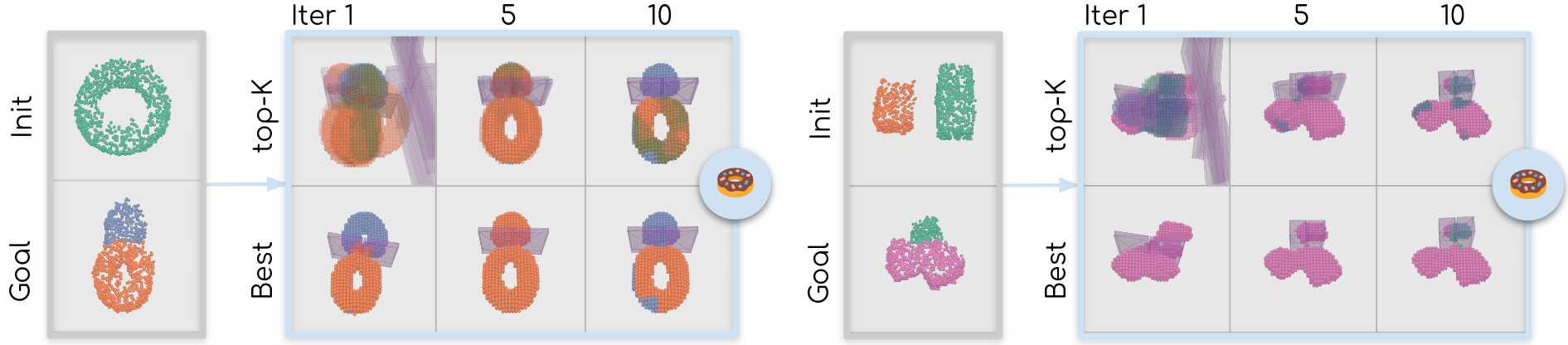}
    \caption{\textbf{Planning Topological Manipulation.} Two examples, each defined by an initial partial observation and a partial goal observation (color-coded topology). We show \ours's predictions for the top-K sampled actions (top) and the one selected based on our latent representation (bottom) after 1, 5, and 10 CEM iterations.}
    \label{fig:planning}
\end{figure}

\subsection{Planning for Topological Manipulation}\label{sec:planning}
We leverage \ours~to plan topological manipulations. Goals may be provided as a mesh or, illustrated in~\cref{fig:planning}, a (partial) point cloud with color-coded components.
Specifically, \ours~is used to roll-out sampled actions in a CEM planner~\cite{deboer2005cem}. We use the cosine similarity between the goal and our predictions in the learned latent space, naturally capturing geometrical and topological fitness for resampling and selection. We initialize $K=64$ actions; in-plane translation and rotation from Gaussian distributions $t_x,t_y\sim\mathcal{N}_t(\vzero , \mI \cdot 8) \in[-40,40]^2$mm and $\theta_z\sim\mathcal{N}_\theta(0,10)\in[-10,10]$deg, as well as the gripper from a multinomial distribution $x\sim\mathcal{M}_x(\text{narrow}=\frac{1}{3},\text{regular}=\frac{1}{3},\text{wide}=\frac{1}{3})$. After each rollout, the sampled actions are scored by similarity, the best $K_{\text{top}}=8$ are kept and used to re-fit the distributions. The best action is selected after ten CEM iterations.
The results in~\cref{fig:planning} visualize the distribution of the top-K and the best samples over multiple CEM iterations, as predicted and reconstructed by \ours. The large initialization variance of in-plane actions and EE geometries quickly reduces and we are able to fulfill both goal aspects, also finding a well-suited EE.
Quantitatively, the action parameters generating the goal observations in~\cref{fig:planning} are accurately estimated by our best plans, achieving translation and rotation errors of 1.6 / 1.1mm and 0.4 / 0.5deg, respectively. Note that this is close to the 1mm grid resolution used for the evaluated reconstructions.


\begin{figure}[!t]
    \centering
    \begin{subfigure}{0.75\textwidth}
        \includegraphics[width=\linewidth]{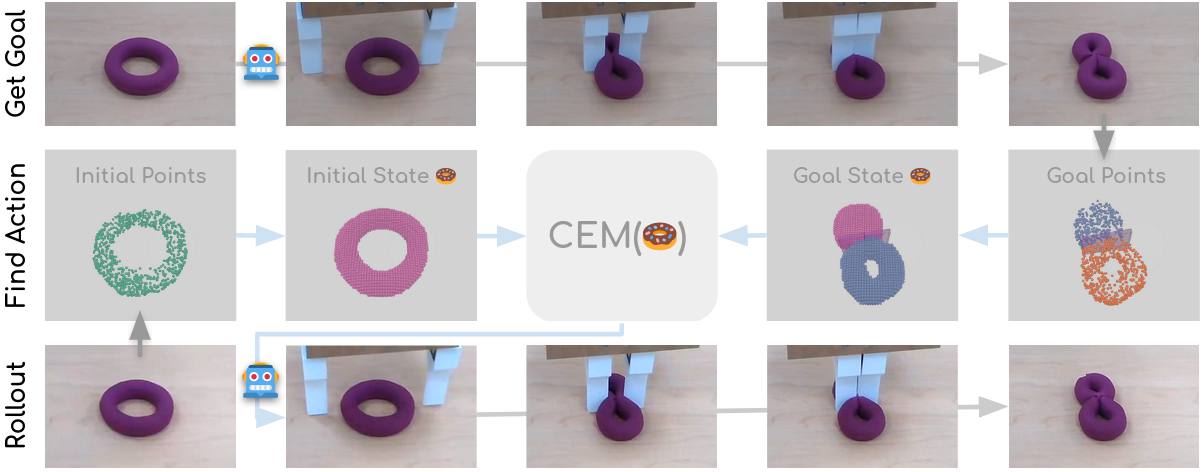}
        \subcaption{\footnotesize Robot-defined goal: a doughnut split by the narrow EE.}
        \vspace{1mm}
        \label{fig:donut-robot-narrow}
    \end{subfigure}
    \begin{subfigure}{0.75\textwidth}
        \includegraphics[width=\linewidth]{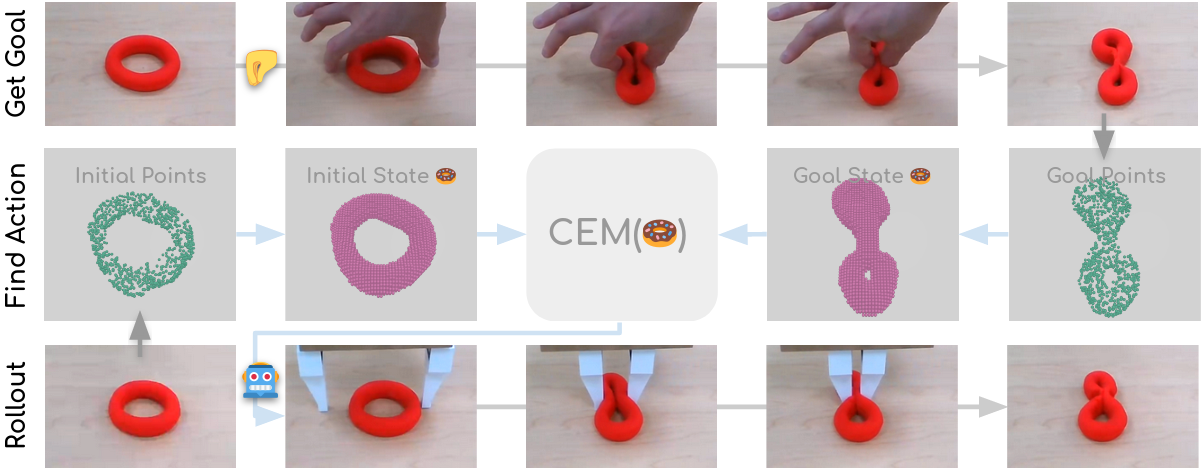}
        \caption{\footnotesize Human-defined goal: a doughnut pinched by fingers. Our approach automatically selects the regular EE ($\sim$ fingers) to achieve the goal.}
        \vspace{1mm}
        \label{fig:donut-human-regular}
    \end{subfigure}
    \begin{subfigure}{0.75\textwidth}
        \includegraphics[width=\linewidth]{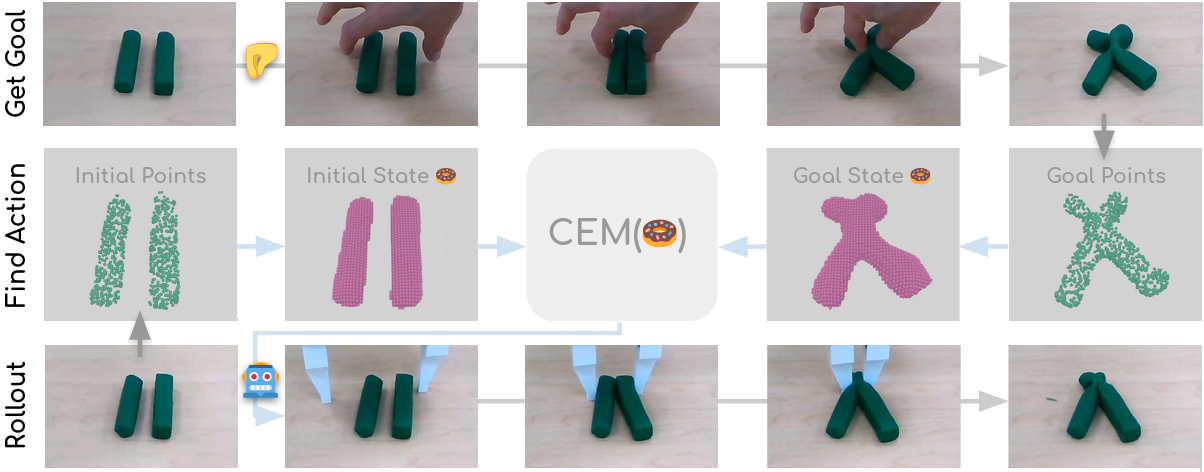}
        \caption{\footnotesize Human-defined goal: two rolls pinched by fingers. Our approach automatically selects the regular EE ($\sim$ fingers) to achieve the goal.}
        \vspace{1mm}
        \label{fig:rolls-human-regular}
    \end{subfigure}
    \begin{subfigure}{0.75\textwidth}
        \includegraphics[width=\linewidth]{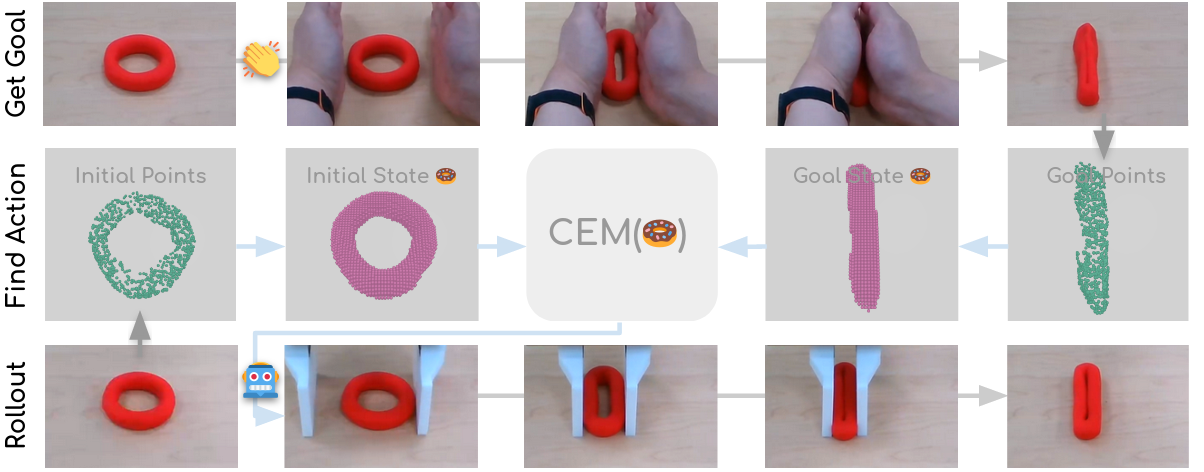}
        \caption{Human-defined goal: two rolls squeezed by palms. Our approach automatically selects the wide EE ($\sim$ palms) to achieve the goal state.}
        \vspace{1mm}
        \label{fig:donut-human-wide}
    \end{subfigure}
    \caption{\textbf{Real-world Planning Results.} Top: Goal point clouds are created by either a robot (a), human fingers (b, c) or human palms (d); manipulating a doughnut (a, b, d) or two rolls (c). The goals are achievable by one of the held-out EE geometries, namely the narrow (a), regular (b, c) or wide EE (d). Middle: DoughNet~\donut~enables a vanilla CEM-based planner to find the best action, given the goal and a partial view of the initial state. Bottom: The found action is executed on a robot, successfully achieving the desired goal state (right-most columns).}
    \label{fig:real_planning}
\end{figure}

\textbf{From Simulated to Real Manipulation.} \ours~ successfully generalizes from synthetic robotic training data to real robotic test cases, as shown in~\cref{fig:donut-robot-narrow}. Given partial point clouds of the goal and initial states, we again employ \ours~to reconstruct complete states and roll-out sampled actions in a CEM-based planner (referred to as \textit{CEM(\donut)} in~\cref{fig:real_planning}). Comparing the predicted latent state to the completed latents of the real goal state, we are able to select the best-suited EE, in-plane pose and final opening width. By executing this plan on the robot, we accurately recreate the desired goal in terms of geometry and topology, as shown in the right-most column in~\cref{fig:donut-robot-narrow}.

\textbf{From Robot-defined to Human-defined Goals.}
Even further from the training data, \ours~enables planning towards real human-defined goals, created by pinching Play-Doh shapes with fingers, or squeezing them with palms. As shown in~\cref{fig:donut-human-regular,fig:rolls-human-regular,fig:donut-human-wide}, our approach is able to identify the correct tool (from held-out EE geometry) to recreate the human manipulation. 
However, while the achieved manipulations are close to the desired goals, we want to also discuss the specific challenges of this transfer to human input. Compared to our synthetic training data, human manipulation creates more irregular object geometries due to the varying \say{manipulator geometry} and its highly non-linear motion (i.e., articulated, deforming fingers and hands). In~\cref{fig:donut-human-regular,fig:rolls-human-regular}, the unequal width of the index finger and thumb, combined with additional movement while pinching, creates an indentation that is too wide for the most-suited of the three held-out EE shapes. Hence, in a best effort, a plausible action that pinches part of that region is found by our approach. Similarly, in~\cref{fig:donut-human-wide}, the curvature of the palms affects the created goal state, whereas our planned robotic manipulation (analogous to the training data) may only create flat contact geometries. Combined with a noisy partial observation, our reconstruction tends to represent a too wide goal state in these cases. In turn, the final opening width is slightly wide; although we note that it is still sufficient to merge the doughnut shape into a single component as desired. 


\subsection{Predicting Topological Change}\label{sec:supp_topology}
Supporting our observations in the ablation study in~\cref{sec:ex_prediction} and~\cref{tab:ablation}, we find our approach to jointly encode geometry and topology in a well-behaved learned latent space; thereby enabling the latent-space supervision and set-to-set prediction discussed in the ablation study.

\begin{figure}[!t]
    \centering
    \includegraphics[width=0.9\linewidth,trim={50mm 80 50 50},clip]{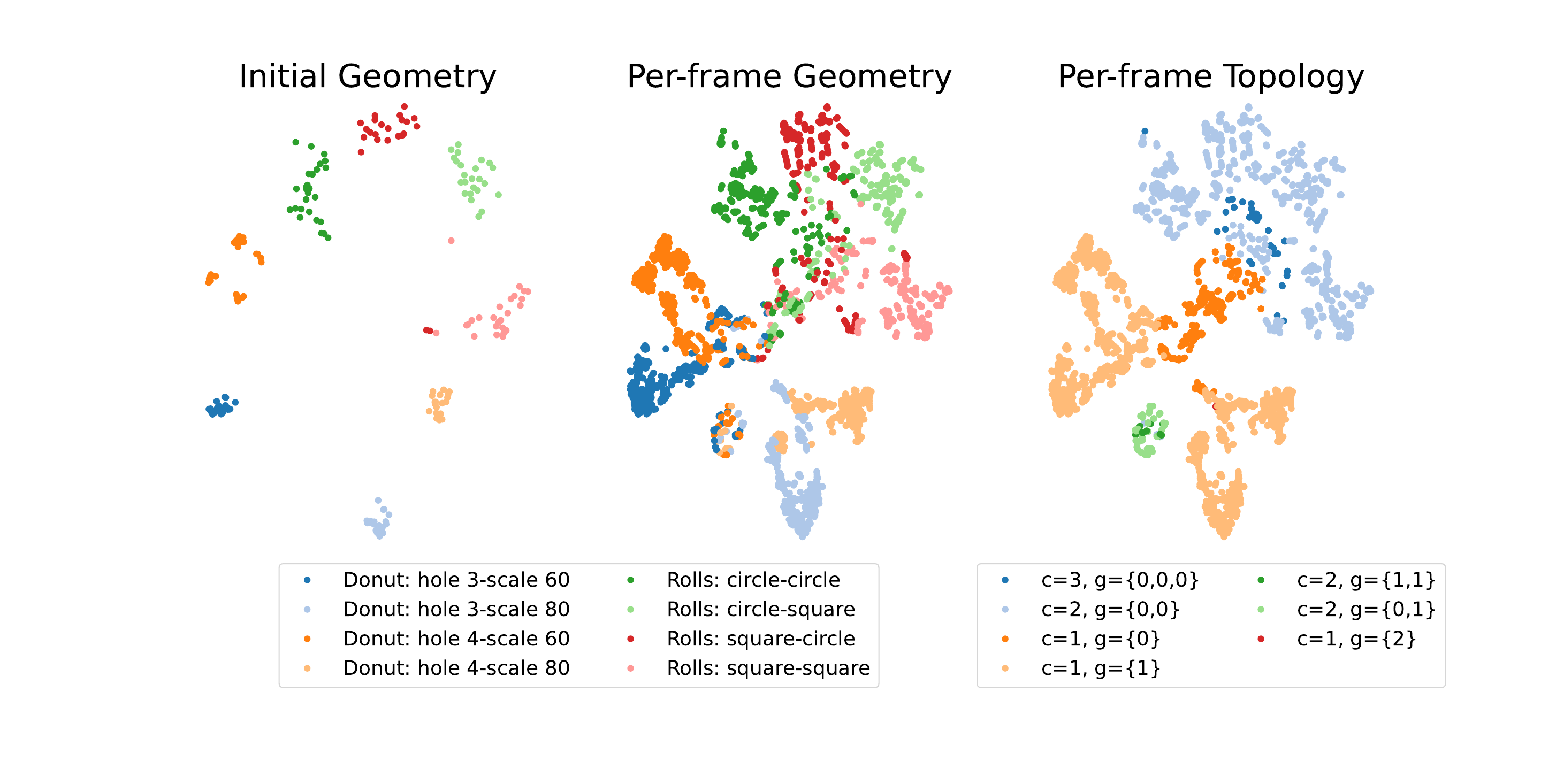}
    \caption{\textbf{t-SNE Visualization.} We show a subset of frames, normalized and reduced to 32 features using PCA before embedding them via t-SNE. Left: The clusters correspond to the initial geometry in the first frame. Middle: Tracking these initial geometries over multiple frames, we observe a regression towards the centroid. Right: Coloring the same clusters by their topology (which may change over time), we find that the regression corresponds to the frequent topological manipulation result of merging to a single component (dark orange). To the top right of this common cluster, we find the topologies that may result from rolls (blue); to the bottom left, the topologies that may result from donuts (light orange, green). We note the equivalent super-cluster boundary for the main types of geometries (donuts and rolls in the left and middle subfigure).}
    \label{fig:tsne}
\end{figure}
\begin{figure}[!t]
    \centering
    \begin{subfigure}{0.37\linewidth}
        \includegraphics[width=\linewidth,trim={0 20 10mm 45},clip]{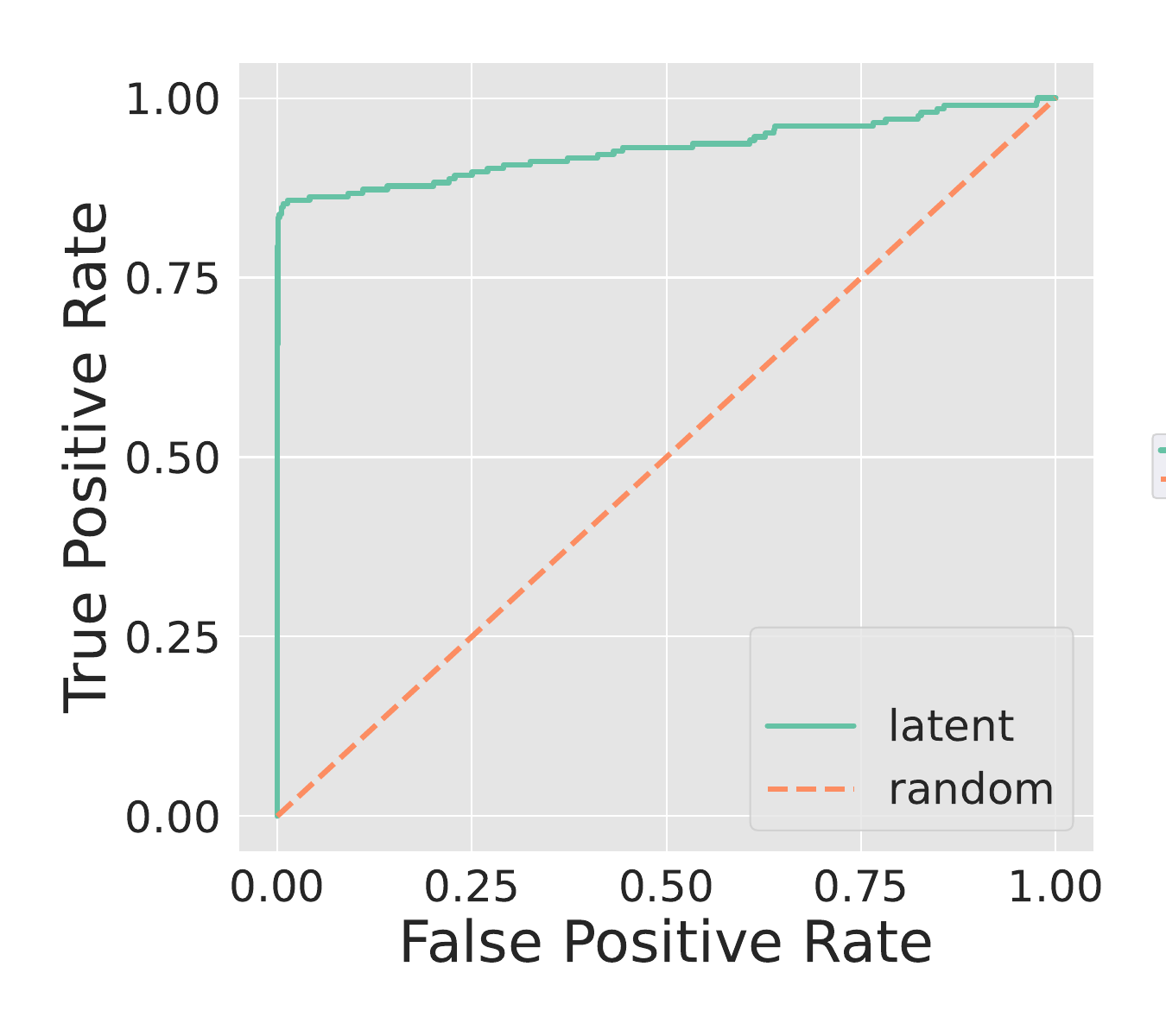}
        \caption{The diagonal line indicates a random binary classifer (\say{change} or \say{no topological change} per frame).}
    \end{subfigure}\hspace{4ex}
    \begin{subfigure}{0.37\linewidth}
        \includegraphics[width=\linewidth,trim={0 20 10mm 45},clip]{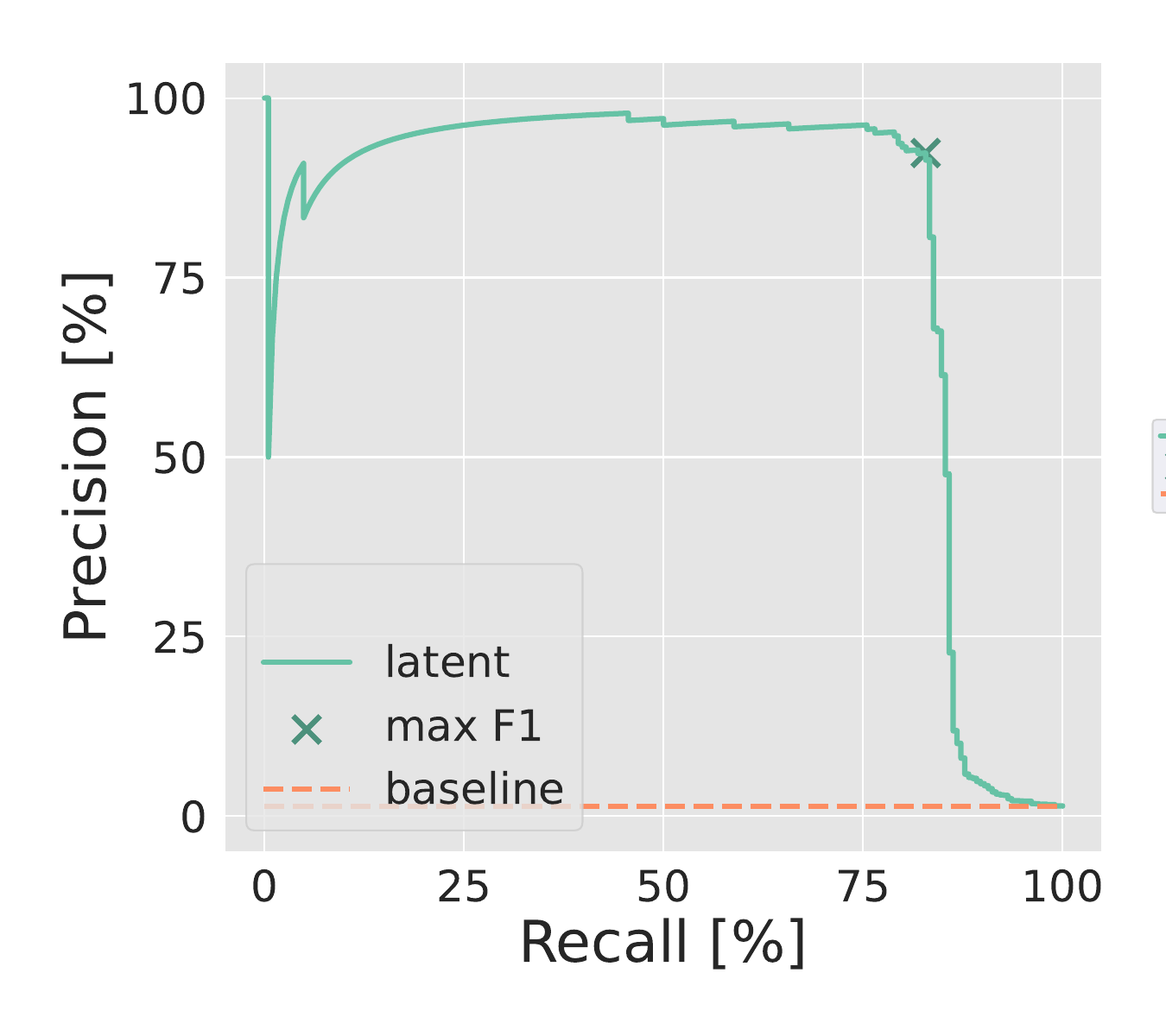}
        \caption{The \textit{baseline} indicates the positive rate of the dataset, i.e., $1.4\%$ of the $75$ raw keyframes (before subsampling).}
    \end{subfigure}
    \caption{\textbf{Predicting Topological Change.} The z-score of the Euclidean distances between subsequent frames in our learned \textit{latent} space allows to predict when a topological change occurs. We achieve an $F_1$-score of $87.3\%$ at a z-score threshold of $4.33$.}
    \label{fig:roc_prec-rec}
\end{figure}

The t-SNE visualization~\cite{van2008tsne} in~\cref{fig:tsne}~(left) shows the different initial geometries in (a subset of) the first frames. Each type of scene is clearly separated; we note that the mirror-equivalent configurations of a roll with circular and a roll with square profile are projected to adjacent regions. Moreover, the super-types (scenes with a donut, and scenes with two rolls) occupy distinct half-spaces in the projection. 
Similarly, although the scene topology changes over subsequent manipulation frames, we observe clusters in~\cref{fig:tsne}~(right) that are consistent with the classes of the resulting topology (in terms of the number of components $c$ and their respective genus $g$).

By tracing individual trajectories, we observe \say{jumps} in the latent space that seem to correlate with topological changes. Based on this, we hypothesize that the distances in the latent space are sufficient to predict whether two subsequent frames relate to different topologies -- that is, if and at which frame such a topological change has occured.
To this end, we define a simple classifier. First, we take the mean over the latent set per frame, leaving a $512$ dimensional vector. We find that the Euclidean distance between two such vectors is discriminative enough of the observed jumps. To detect such \say{outliers} in the frame-to-frame distance, we compute the z-score using the statistic for a sequence of frames and evaluate the resulting classifier for varying thresholds on the z-score. \cref{fig:roc_prec-rec}~shows the resulting false-positive and true-positive rates in a ROC curve, as well as the corresponding precision-recall curve. Based on this, we determine the ideal threshold by selecting the one that maximizes the $F_1$ score, achieving an accuracy of $87.3\%$. This experiment illustrates the \say{topology awareness} of our learned latent space, sufficing a distance-based classifier to predict at which time a topological change happens in an observed manipulation sequence.

\subsection{Limitation and Failure Case}
\begin{wrapfigure}{r}{0.5\linewidth}
    \vspace{-5mm}
    \centering
    \includegraphics[width=0.5\textwidth]{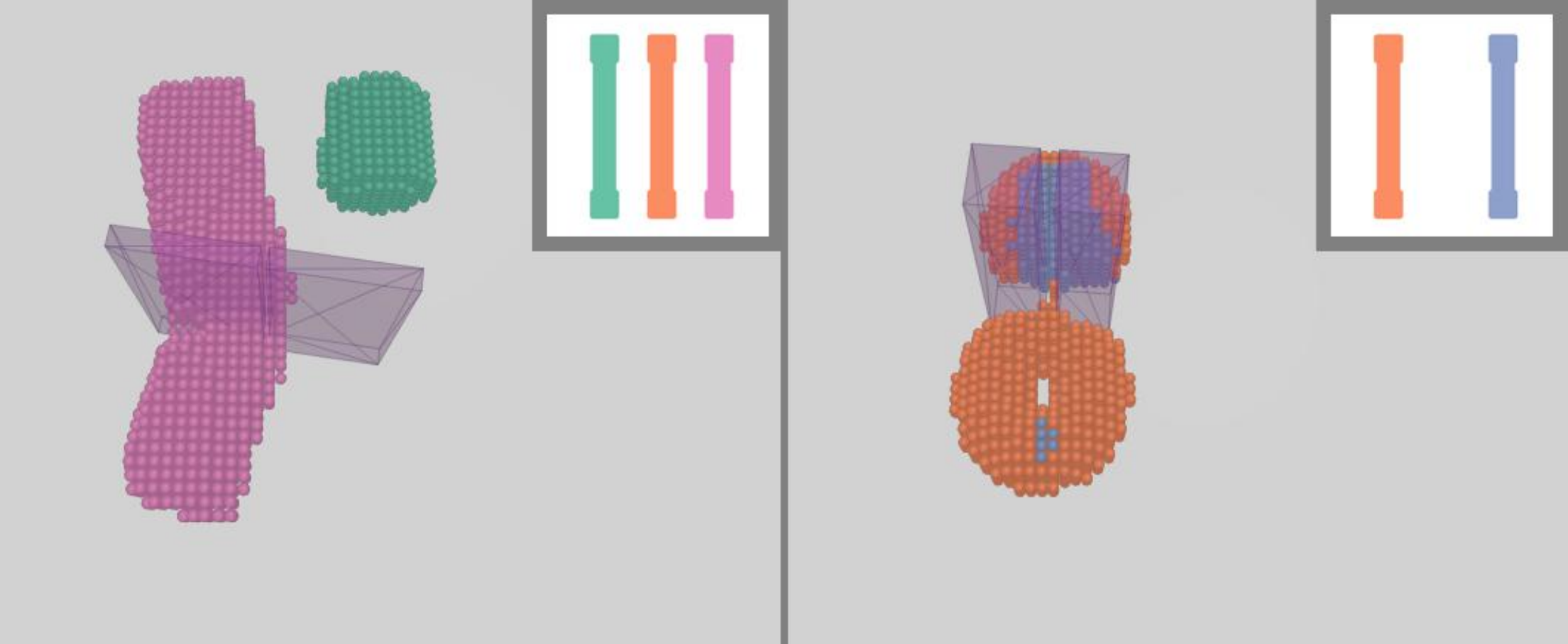}
    \caption{\footnotesize \textbf{Failure Cases.} Inconsistency between occupancy and topology (left) and within changing components (right) of the predicted object state.}
    \label{fig:failure}
\end{wrapfigure}

As shown in~\cref{fig:failure}, we observe two common failure cases in \ours's predictions. First, since this is not enforced, the predicted component occupancy and topology may be inconsistent (two geometrical components but three topological ones predicted in the left example). This is related to a tendency to delayed, or even missed, splitting events. Second, topological changes may result in inconsistency within the occupancy prediction. In the right example, the new components are still partially assigned to one another. However, we observe that this mistake is typically resolved in subsequent steps, presumably as such erroneous shapes (and hence their latents) are mapped back to the learned manifold of shapes observed during training.

Stemming from our dataset design, targeted at providing a minimal test bed for topological manipulation, we only consider individual squeezing actions and a single set of material properties. While these added complexities are beyond the scope of this work, we hope that the release of our data generator will facilitate future work on topological manipulation.

Finally, we do not anticipate any negative societal impact of this work, if responsibly incorporated in robotic systems such that potentially unreliable or erroneous predictions may be considered via verification before open-loop execution, or via reinitialization of short prediction horizons in closed-loop control.

\section{Conclusion}
\label{sec:conclusion}
We propose~\textbf{\ours}, a visual predictive model that jointly reasons about geometrical deformation and resulting topological changes. In our experiments using simulated and real robotics interactions, our approach accurately predicts splitting and merging of deformable objects, manipulated by varying end-effector shapes. In addition, we propose a synthetic data generator to facilitate further research into \textit{topological manipulation}, where geometrical and topological goals may constrain each other.

The successful sim-to-real transfer of our experimental setup should motivate future work extending it towards more complex, sequential topological manipulation tasks. As a predictive model, \ours~lends itself to be employed in robotic planning pipelines; its general notion of interactions via a tuple of \say{manipulator} geometries moreover extends its application possibilities beyond robotics, while its speed and differentiability opens up integration into existing end-to-end trained or optimization-based approaches.

\vspace{3ex}
\noindent{\large\bf Acknowledgements}
\vspace{3ex}

This work was supported in part by the Toyota Research Institute, NSF Award \#2143601, and Microsoft Fellowship. We would like to thank Google for the UR5 robot hardware. Dominik Bauer is partially supported by the Austrian Science Fund (FWF) under project No. J 4683. The views and conclusions contained herein are those of the authors and should not be interpreted as necessarily representing the official policies, either expressed or implied, of the sponsors.

\bibliographystyle{plainnat}
\bibliography{ms}
\end{document}